\documentclass[10pt]{article}
\usepackage[utf8]{inputenc}
\usepackage[letterpaper, margin=1in]{geometry}
\usepackage{cite}
\usepackage{amsmath}
\usepackage{bbm}
\usepackage{graphicx}
\usepackage[caption=false,font=footnotesize]{subfig}
\usepackage{lscape}
\usepackage{caption}
\providecommand{\keywords}[1]{\textbf{Keywords---} #1}

\title{\textbf{Benchmarking features from different radiomics toolkits / toolboxes using Image Biomarkers Standardization Initiative}}
\author{Mingxi Lei$^1$, Bino Varghese$^2$, Darryl Hwang$^2$, Steven Cen$^2$,\\ Xiaomeng Lei$^2$, Afshin Azadikhah$^2$, Bhushan Desai$^2$,\\ Assad Oberai$^3$ and Vinay Duddalwar$^2$}
\date{%
    $^1$Ming Hsieh Department of Electrical Engineering, University of Southern California, 3740 McClintock Avenue, Los Angeles, CA, 90089-2560, USA\\%
    $^2$Department of Radiology, University of Southern California, 1500 San Pablo Street, Los Angeles, CA 90033, USA\\
    $^3$Department of Aerospace and Mechanical Engineering, University of Southern California, 854 Downey Way, Los Angeles, CA 90089, USA\\[2ex]
}

\begin{document}
\captionsetup[figure]{labelfont={bf},name={Fig.},labelsep=space}
\captionsetup[table]{labelfont={bf},name={Table},labelsep=space}
\maketitle

\begin{abstract}
\normalsize
There is no consensus regarding the radiomic feature terminology, the underlying mathematics, or their implementation. This creates a scenario where features extracted using different toolboxes could not be used to build or validate the same model leading to a non-generalization of radiomic results. In this study, the image biomarker standardization initiative (IBSI) established phantom and benchmark values were used to compare the variation of the radiomic features while using 6 publicly available software programs and 1 in-house radiomics pipeline. All IBSI-standardized features (11 classes, 173 in total) were extracted. The relative differences between the extracted feature values from the different software and the IBSI benchmark values were calculated to measure the inter-software agreement. To better understand the variations, features are further grouped into 3 categories according to their properties: 1) morphology, 2) statistic/histogram and 3)texture features. While a good agreement was observed for a majority of radiomics features across the various programs, relatively poor agreement was observed for morphology features. Significant differences were also found in programs that use different gray level discretization approaches. Since these programs do not include all IBSI features, the level of quantitative assessment for each category was analyzed using Venn and the UpSet diagrams and also quantified using two ad hoc metrics. Morphology features earns lowest scores for both metrics, indicating that morphological features are not consistently evaluated among software programs. We conclude that radiomic features calculated using different software programs may not be identical and reliable. Further studies are needed to standardize the workflow of radiomic feature extraction.
\end{abstract}

\section{Introduction}
Radiomics, has recently been the focus of a lot of interest and research as a method of extracting more information from clinical imaging and therefore help solve different clinical problems. With the advances in image processing and data science, medical images (computed tomography, magnetic resonance, positron emission tomography images, etc.) are converted to minable high-dimensional data, which may carry latent information that can facilitate clinical decision-making. In general, a radiomics pipeline consists  of 4 steps: image acquisition, tumor segmentation, radiomics feature extraction and statistical/machine learning-based analysis \cite{Court2016}. A number of studies have shown that any of these 4 steps can cause the variability of radiomics features \cite{Lee2019,Monica2018,Bianchi2019,Foy2018,Shafig2017,Bogowicz2016,Kumar2012,Schwier2019}. For example, currently, different imaging centers use varying image acquisition protocols. The lack of consensus or guidelines on image acquisition across institutions leads to imaging data heterogeneity among different studies \cite{Mackin2015}. The accuracy of manual segmentation also affects the robustness and stability of radiomics models, however it is a viable solution in scenarios of contouring highly varying complex tumor boundaries with low tumor-tissue contrast using automated methods \cite{vanVelden2016}.

While researchers have recognized the relative lack of consistency and reproducibility, there is a dearth of approaches to address this issue.  Many research institutions have developed their in-house or open-source programs for radiomic analysis. These programs usually run different implementations of the same features but with different terminologies. In some cases, the same metrics may not generate identical values due to differences in implementing the underlying algorithms e.g. image intensity normalization based on just the tumor region or on entire image may generate different results. 

To address the standardization of radiomics features, the image biomarker standardization initiative (IBSI), an international collaboration, was founded \cite{Zwanenburg2020}. The IBSI have fully propose the definitions of 11 commonly used feature classes. A digital phantom and a CT phantom, with their benchmark values of features, have been created for the purpose of standardization. 

Recent analyses have been conducted on datasets for evaluating the variation of features among radiomics software programs, showing that large variations were found due to image preprocessing, parameters and algorithm implementations \cite{Monica2018,Bianchi2019,Foy2018}. As these studies focused on specific private datasets and limited number of features, the results are not generalizable. In this study, the issue of generalization was addressed by using the IBSI benchmarking to generate reliable results.

\section{Materials and Methods}
\label{sec:method}
\subsection{Digital Phantom}
\label{sec:method1}

\begin{figure}
    \centering
    \includegraphics[width=84mm]{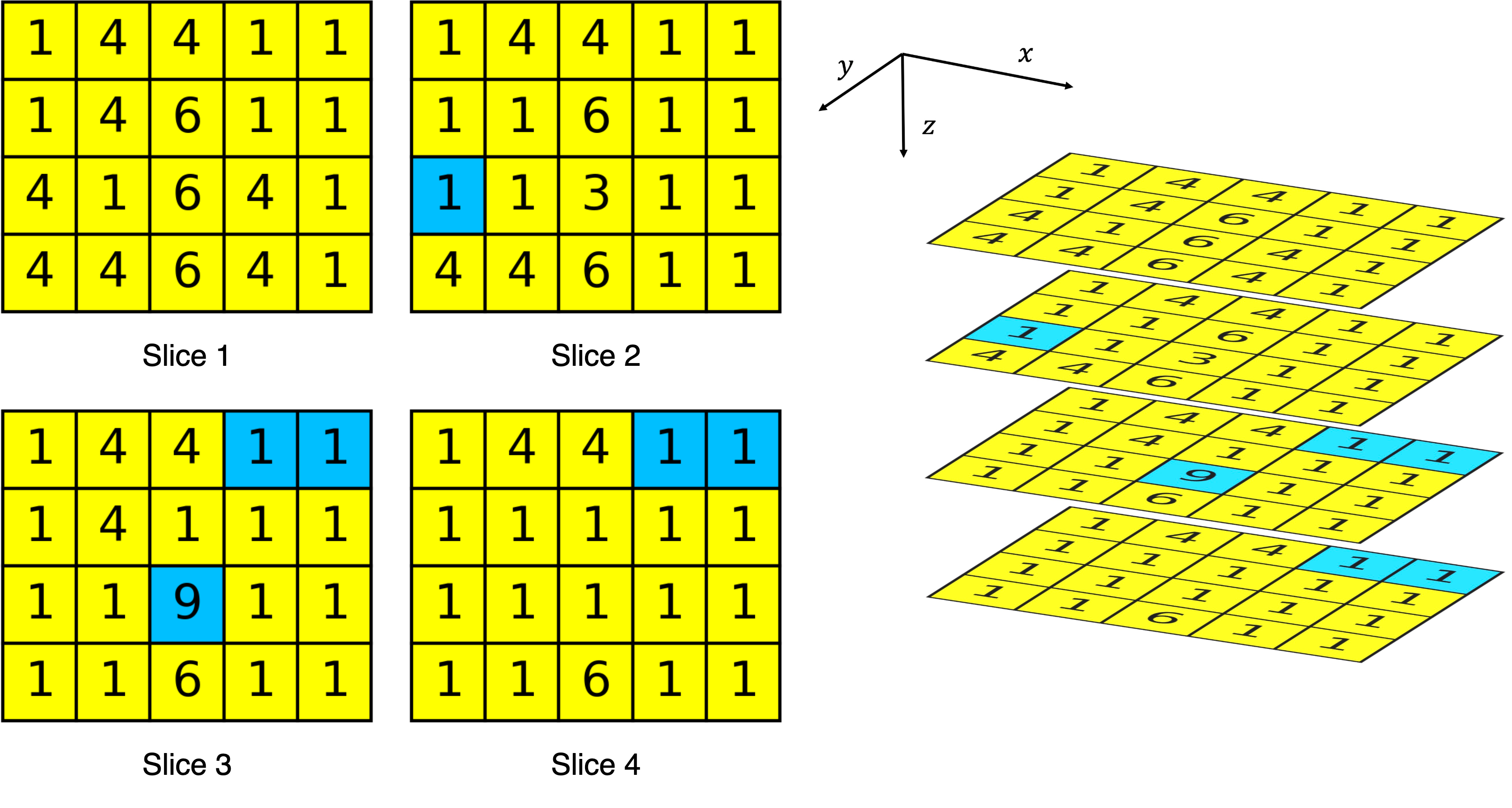}
    \caption{IBSI Digital Phantom. The grey levels are annotated in each voxel. Yellow region indicates the ROI}
    \label{fig:1}
\end{figure}

The digital phantom developed by the IBSI is geometrically small, consisting of 5 $\times$ 4 $\times$ 4 voxels. Fig. \ref{fig:1} shows four slices of the phantom. Grey levels are designed as integers and hence, discretization is not required before computing features. Grey levels range from 1 to 9 in the whole volume and from 1 to 6 in the region of interest (ROI). Level 2 and 5 are absent. Voxels are isotropic, with 2.0 $\times$ 2.0 $\times$ 2.0 mm spacing for the three dimentions.
\subsection{IBSI-standardized Features}
To comprehensively compare inter-software differences, 11 classes of radiomics features were evaluated in this study: morphology, local intensity, intensity-based statistic, intensity histogram, intensity-volume histogram, grey level co-occurrence matrix (GLCM), grey level run length matrix (GLRLM), grey level size zone matrix (GLSZM), grey level distance zone matrix (GLDZM), neighborhood grey tone difference matrix (NGTDM) and neighboring grey level dependence matrix (NGLDM) based features. These features are standardized and summarized from various technical references in the IBSI. According to their mathematical definitions, the 11 classes of radiomics features characterize various properties of ROI, and they can be grouped into 3 major categories: 1) statistical/histogram, 2) texture, and 3) morphology features (Table \ref{tab:1}).

Table \ref{tab:2} shows the number of features within the IBSI standards that fall within each of the aforementioned categories. While the IBSI has a specified terminology, similar features may be named differently in different software programs. As an example, GLCM joint maximum is tagged as maximum probability in some programs e.g. Pyradiomics \cite{van_Griethuysene2017}, QIFE \cite{Echegaray2018} or studies \cite{Aerts2014,Wu2016,COROLLER2015345}. In some software programs \cite{van_Griethuysene2017}
, kurtosis is computed with the formula of excess kurtosis, which is  an alternative definition of kurtosis. While excess kurtosis is equal to kurtosis – 3, to perform comparisons to IBSI standards the values of kurtosis must be corrected manually after running these software programs. To conduct a fair comparison of metrics across the different software, we created a glossary of matching radiomic metrics, with their different terminologies, across different radiomics software (Supplementary 1).

\subsection{Radiomics Software Programs}
Six publicly available software programs: Pyradiomics \cite{van_Griethuysene2017,Dou2018,vanGriethuysen2020}, the Medical Imaging Interaction Toolkit (MITK) \cite{GOTZ2019108,Kickingereder5765,Kickingereder2016}, LIFEx \cite{Nioche,Nioche01052017,Nardone2018}, the Standardized Environment for Radiomics Analysis (SERA) \cite{Ashrafinia2019,Ashrafinia01052018,Du2020}, Cancer Imaging Phenomics Toolkit (CaPTk) \cite{Davatzikos2018,Pati2020,Rathore2018}, a MATLAB library from McGill University (A2, GitHub Repo: radiomics-develop) \cite{Zwanenburg2020,Valli_res_2015,Vallieres2017} and an in-house MATLAB library (A1, University of Southern California \cite{Hassani2019,Varghese2019,Varghese20192}) are evaluated here (Table \ref{tab:3}). These programs were included in our study as they are widely used in radiomics research and span across different software development languages and operating systems.

While the underlying algorithm remains the same, its application differs across different software (Table \ref{tab:4}). For example, for morphology features, researchers commonly use Marching Cubes algorithms for meshing presentation, as it works efficiently on multiple programming platforms \cite{Lorensen1987}. However, more efficient implementations were proposed by other researchers \cite{NEWMAN2006854}. In fact, the IBSI also recommend the methodology proposed by Lewiner \cite{Lewiner2003}, optimizing the performance and feasibility. Also, intensity-based statistics and histogram features are already well defined in common image processing field and therefore, their computing results should mainly rely on relatively established programming strategies. For example, for textural features, all software basically cite the same publications, where GLCM was proposed by Haralick \cite{Haralick1973}, GLRLM was proposed by Galloway \cite{GALLOWAY1975172}, GLSZM was proposed by Thibault \cite{Thibault2009,Thibault2014}, GLDZM was derived from Thibault \cite{Thibault2014}, NGTDM was proposed by Amadasun \cite{Amadasun1989} and NGLDM was proposed by Sun \cite{SUN1983341}.

\begin{figure}
    \centering
    \subfloat[]{\includegraphics[width=84mm]{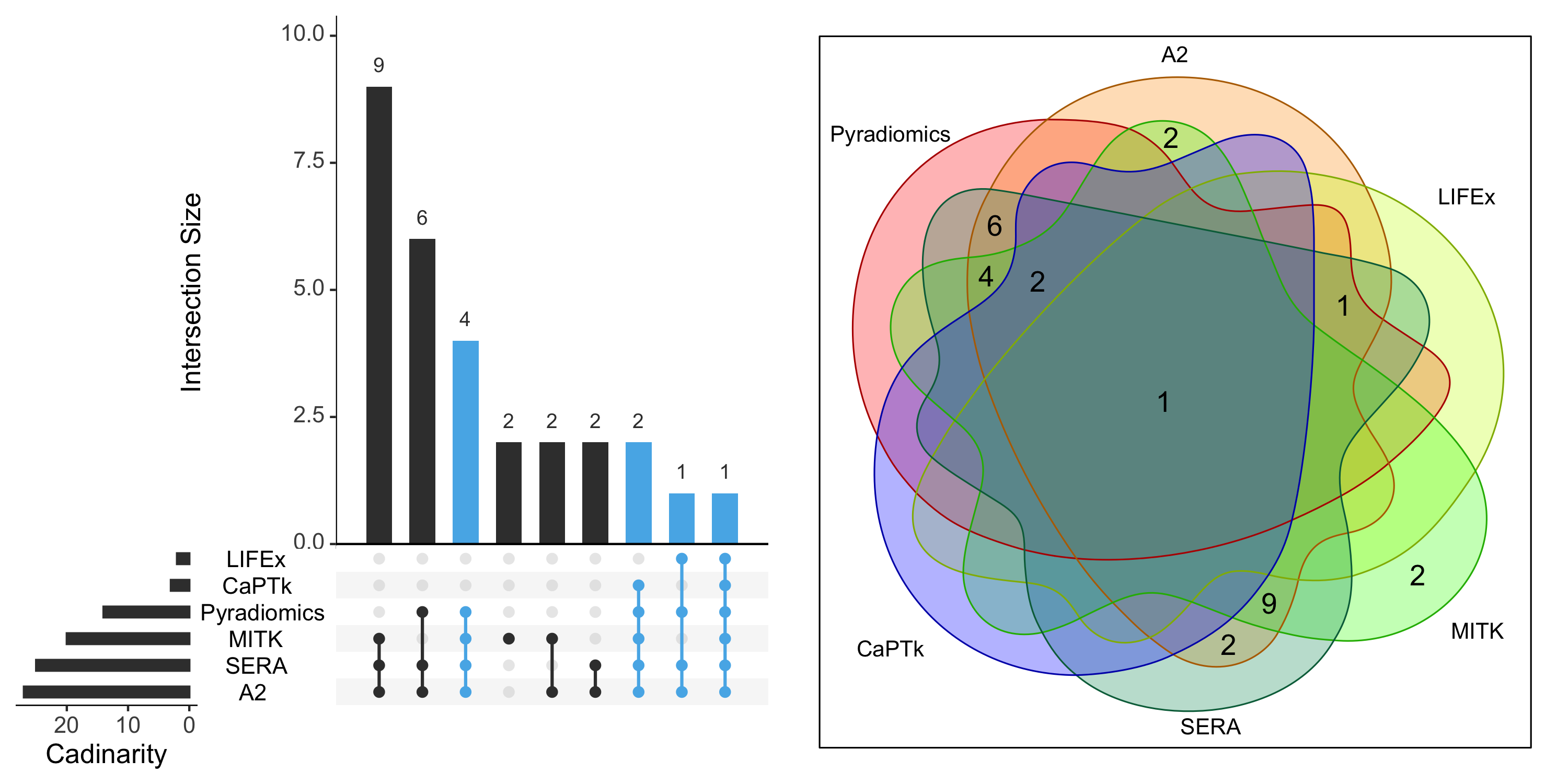}\label{fig:2-1}}
    \hfil
    \subfloat[]{\includegraphics[width=84mm]{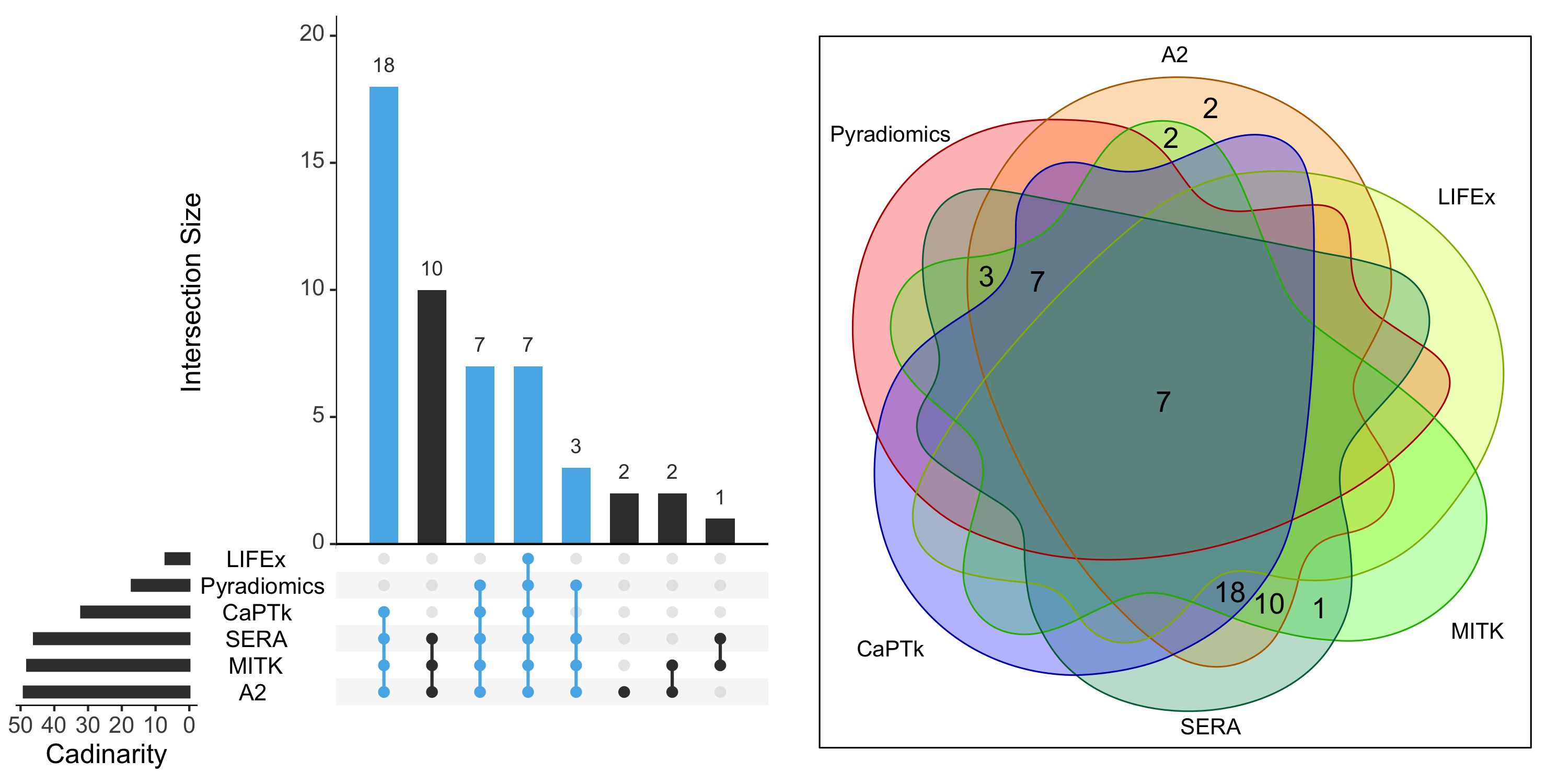}\label{fig:2-2}}
    \hfil
    \subfloat[]{\includegraphics[width=84mm]{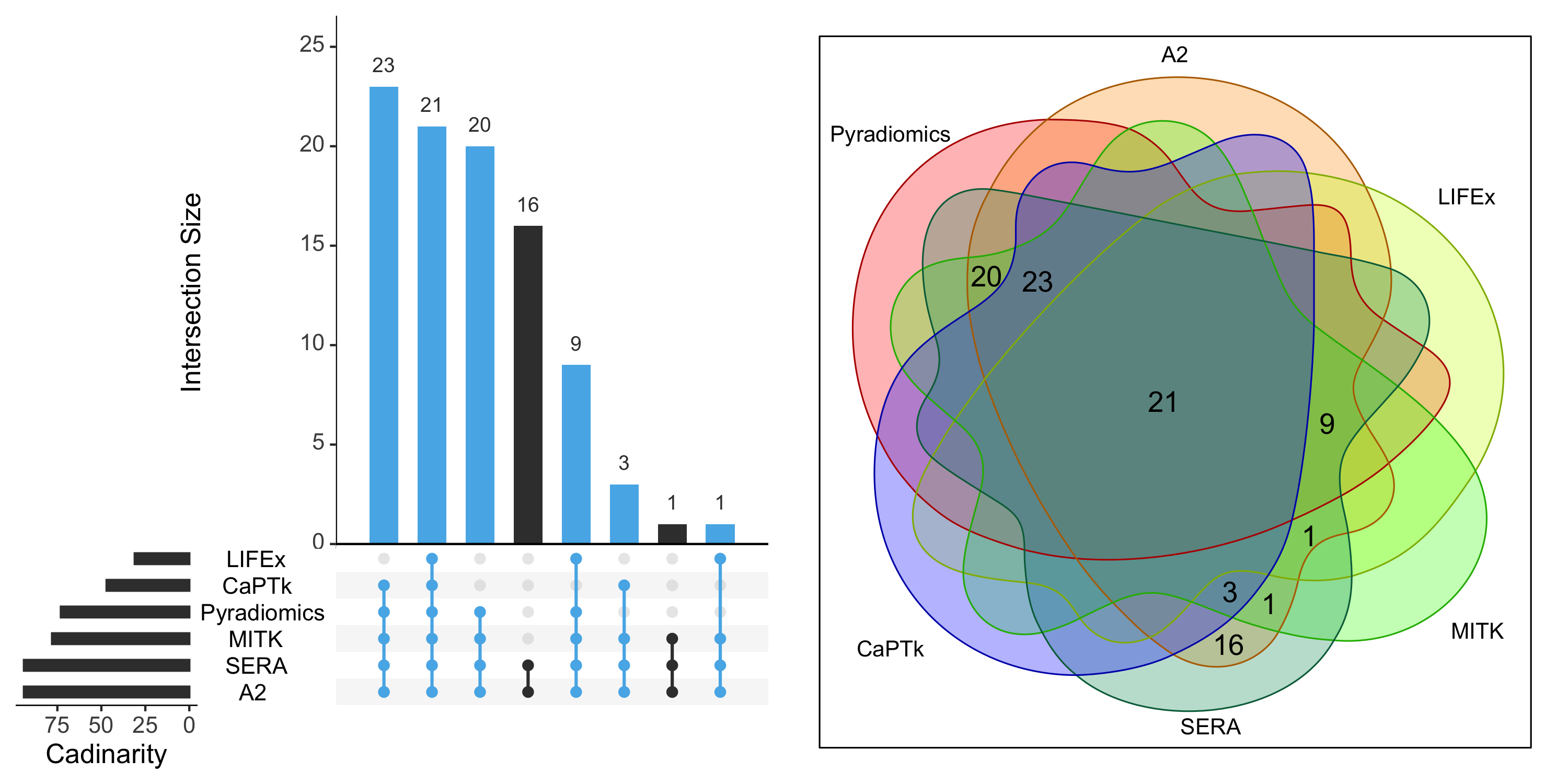}\label{fig:2-3}}
    \hfil
    \subfloat[]{\includegraphics[width=84mm]{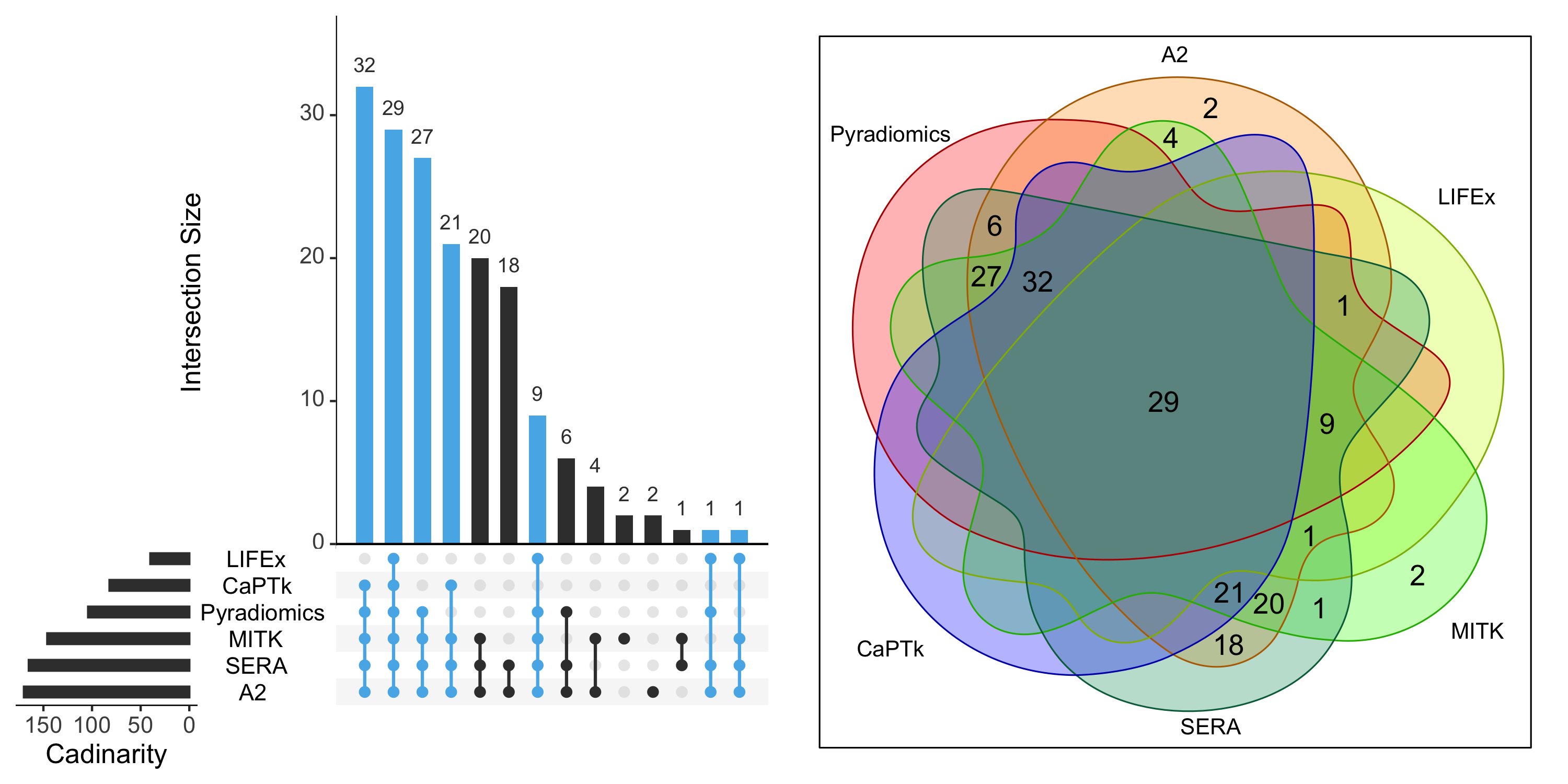}\label{fig:2-4}}    
    \caption{the Venn and UpSet Diagram for Radiomics Features across 6 publicly available programs. (a) morphology features. (b) statistics and histogram features (c) texture features, (d) all IBSI features}
    \label{fig:2}
\end{figure}

\begin{figure}
    \centering
    \includegraphics[width=84mm]{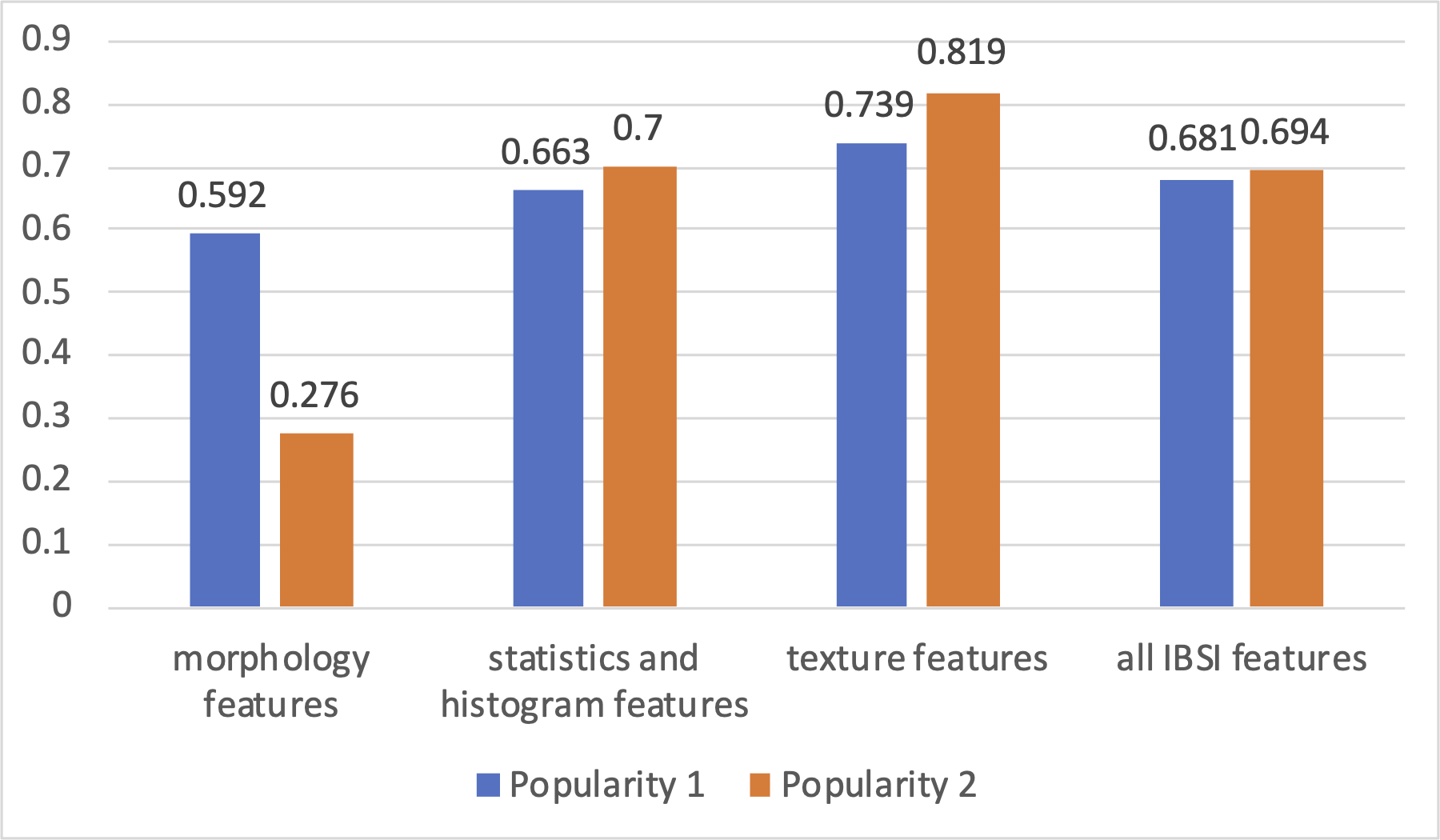}
    \caption{Popularity 1 and Popularity 2 within 3 categories and within all 173 IBSI features}
    \label{fig:3}
\end{figure}

The level of quantitative assessment provided by different radiomic software is different, i.e., not all software programs includes all 173 IBSI-standardized features. Table \ref{tab:1} shows the number of features computed by each software. To better understand the distribution of the 173 different IBSI-defined features across the different radiomics software, an UpSet diagram and Venn diagram (Fig. \ref{fig:2}) approach was used. Using this approach, the “popularity” of the 3 major categories of features were analyzed, i.e., how many features were shared by different radiomics software programs was identified. Only the 6 publicly available software programs were considered in the diagrams for the reason that they may be designed for general radiomics analysis, while in-house software may be designed for application-specific scenarios. As the total number of features of each category are different across and within software, two ad hoc metrics to quantify the “popularity” of the each of the 3-feature category, based on the 6 software were defined:
\begin{equation}
\label{eq:1}
Popularity~1: P_1 = \frac{\sum^d_{i=1}w_i}{6d}, 0\leq P_1 \leq 1
\end{equation}
\begin{equation}
\label{eq:2}
Popularity~2: P_2 = \frac{\sum^d_{i=1}\mathbbm{1}(w_i>4)}{d}, 0\leq P_2 \leq 1
\end{equation}
where $w_i$ indicates the number of software that support the $i^{th}$ features, $d$ indicates the number of features. Features supported in more than 4 out of 6 software programs are considered as "popular". These are highlighted in the UpSet diagram in Fig. \ref{fig:2}. Intuitively, $P_1$ quantifies the weighted cardinality of the specific group of radiomics features; $P_2$ quantifies the percentage of "popular" features in the specific group. Large values in both metrics indicate that radiomics features are widely supported by various software (Fig. \ref{fig:3}). 

\subsection{Computational Workflow}
\begin{figure}
    \centering
    \subfloat[]{\includegraphics[width=84mm]{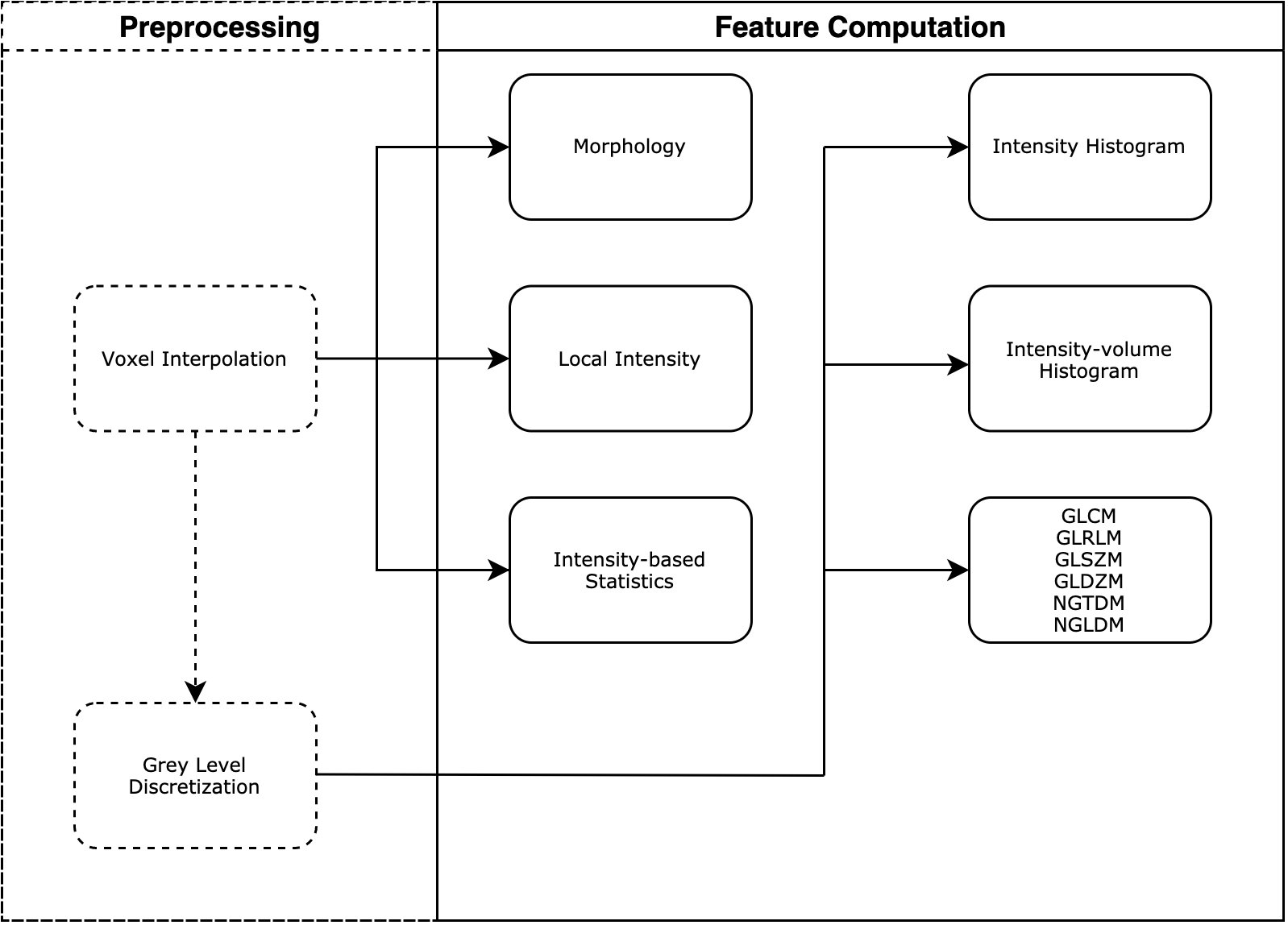}\label{fig:4-1}}
    \hfil
    \subfloat[]{\includegraphics[width=84mm]{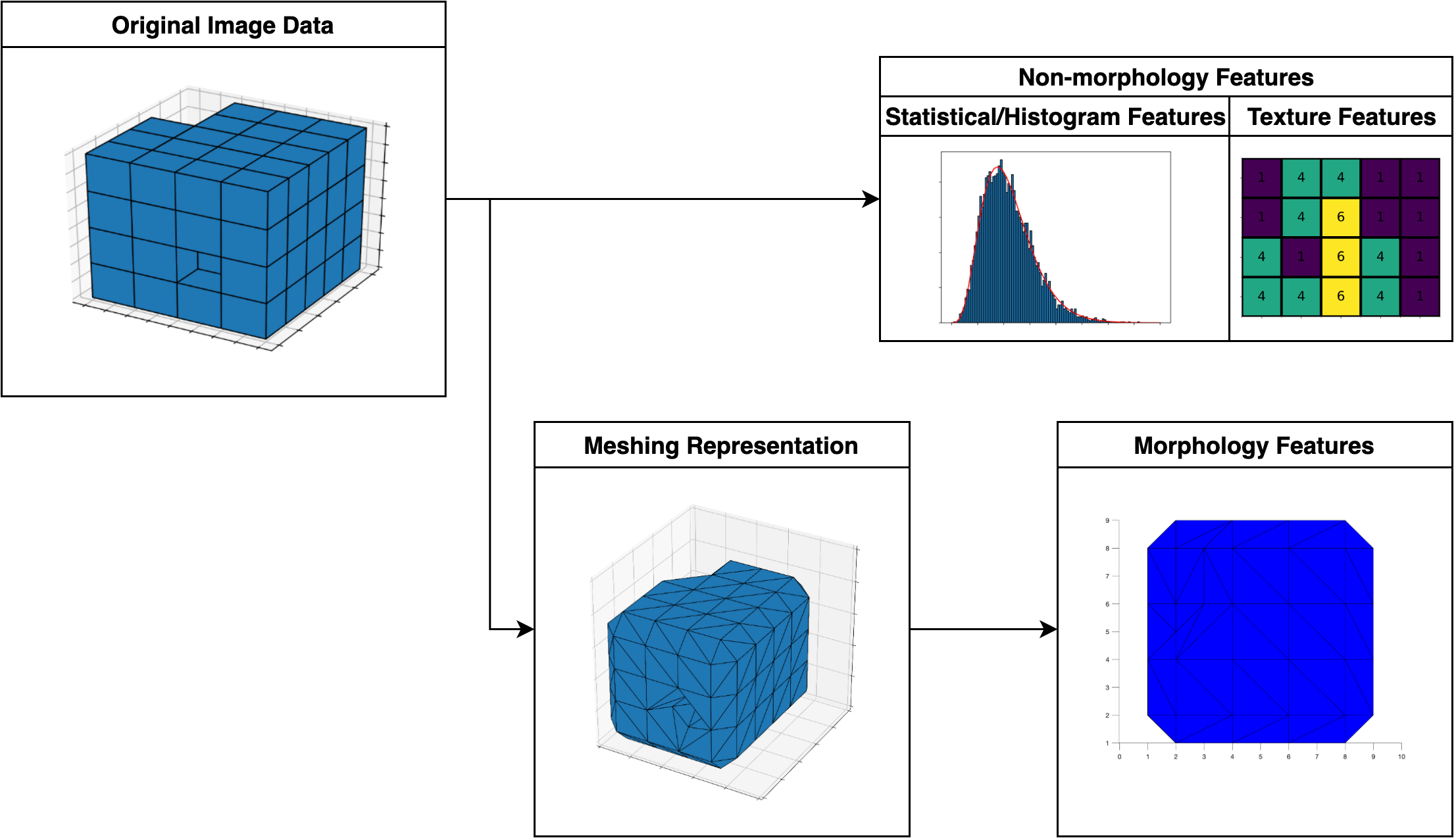}\label{fig:4-2}}
    \caption{(a) General computation workflow. Preprocessing is not required but irremovable in CaPTk and LIFEx, thus enclosed by dash box. (b) Different workflow for morphology features and non-morphology features}
    \label{fig:4}       
\end{figure}

In general, the computational workflow can be divided into two stages: 1) preprocessing and 2) feature computation (Fig. \ref{fig:4-1}). Two preprocessing operations are included in most studies and introduced in the IBSI: voxel interpolation and intensity discretization.

As texture features are sensitive to voxel size variation, interpolation is required. Different algorithms for interpolation, such as nearest neighbor, bilinear, B-Spline interpolation are used. 
Intensity discretization is useful for producing a reasonable quantitative analysis and removing noise. Two approaches are defined in the IBSI: 1) fixed bin number and fixed bin width \cite{Zwanenburg2020}:
\begin{equation}
\label{eq:3}
  X_{d,k} =
    \begin{cases}
      \Big \lfloor N_g \frac{X_{gl,k}-X_{gl,min}}{X_{gl,max}-X_{gl,min}} \Big \rfloor +1 & X_{gl,k} < X_{gl,max}\\
      Ng & X_{gl,k} = X_{gl,max}\\
    \end{cases}
\end{equation}
\begin{equation}
\label{eq:4}
    X_{d,k}=\Big \lfloor \frac{X_{gl,k}-X_{gl,min}}{w_b} \Big \rfloor + 1
\end{equation}
where $X_{d,k}$ denotes the discrete intensity for voxel $k$, $N_g$ denotes the number of bins, $X_{gl,k}$ denotes the original gray level of voxel $k$, $X_{gl,min}$ and $X_{gl,max}$ denotes the minimum and maximum gray level in the ROI, and $w_b$ denotes the interval width of each bin. Number of bins $N_g$ and width of bins $w_b$ need to be specify for fixed bin number and fixed bin width, respectively.

As the digital phantom was developed using discretized intensities and isotropic voxels, no preprocessing is needed. While most of the radiomics software programs allow modular computing, LIFEx and CaPTk are rigid and preprocessing cannot be skipped. To address this problem, specific parameters of voxel interpolation and intensity discretization are assigned to remove their respective contribution. Expected spacing size is needed to specify for interpolation. Spacing size 2.0 $\times$ 2.0 $\times$ 2.0 mm, which is identical to the original spacing, is set for voxel interpolation because we expect the spacing not to change. For intensity discretization, LIFEx use fixed bin width approach with $w_b$ equal to 1 and CaPTk use fixed bin number approach with $N_g$ equal to 6. Similar to the setting for interpolation, we expect both of these settings for discretization do not change voxel intensities. In this study, five basic statistic features are chosen to compare mean, median, minimum, maximum discretized intensity and discretized intensity range across the different software, after intensity discretization to reveal basic properties of intensity distribution which has been used to evaluate the different software. 

Feature computation is the next step after either voxel interpolation or intensity discretization (Fig. \ref{fig:4-1}). For non-morphology features, the feature values are directly calculated from the 3D volume data. Morphology features, however, by definition, are defined under polygon meshing representation, leading to a two-stage computation workflow (Fig. \ref{fig:4-2}). Image volume data is first transformed to meshing representation, then morphology features are calculated based on it.

Table \ref{tab:5} showsshows the parameters in each step of computation. To minimize the effect of different parameters and generalize the results, parameter settings are designed to maintain the highest consistency. For GLCM, GLRLM, GLSZM, GLDZM, NGTDM and NGLDM features, voxel neighborhoods are measured along all possible direction/angle in 3-Dimensional (3D) space, which mean that a central voxel has 26 neighbors. The distance between the central voxel and the neighbor voxel is set as 1 by default. As each direction can generate one GLCM and GLRLM by definition, feature aggregation should be used to summarize feature values from all directions. In general, the information of features from multiple directions are summarized in two different ways. LIFEx averages the features over 26 directions, while the other software programs compute the features after merging matrices from all directions.

\section{Results}
As the voxel interpolation and intensity discretization are fixed in CaPTk and LIFEx, we compare the agreement of these 5 features first to confirm that preprocessing does not change the voxel intensities (Table \ref{tab:6}). Values of mean, median, minimum and maximum intensity calculated by CaPTk are 1 unit less than the benchmark values, while the intensity range is equal. This indicates that CaPTk set the minimum grey level as 0 instead of 1 and all voxel intensities are shifted by -1. LIFEx does not include the 5 features, hence effect of preprocessing is unclear.

\begin{figure}
    \centering
    \includegraphics[width=129mm]{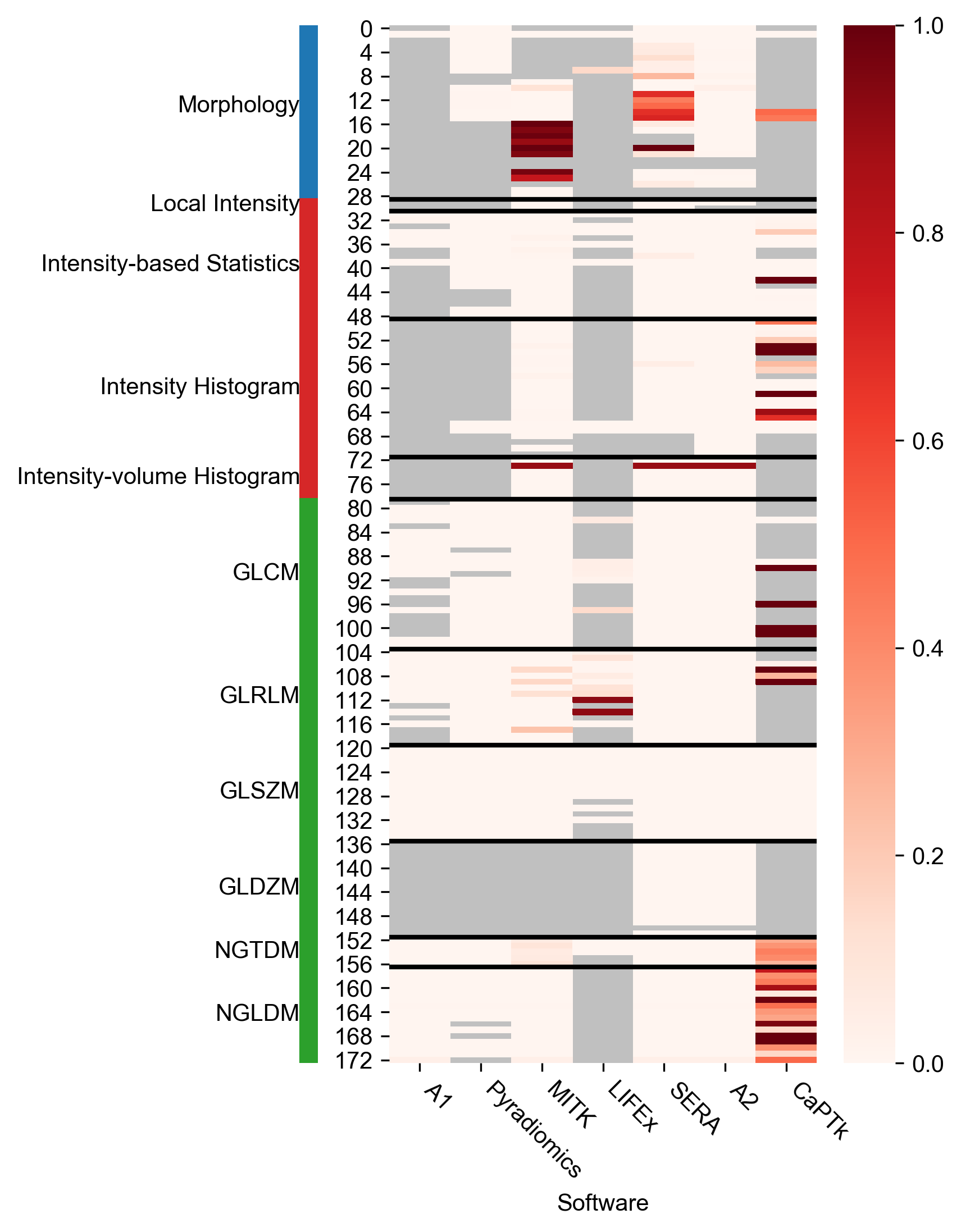}
    \caption{Heatmap for the relative differences between computed features and the IBSI benchmark values. Missing features are denoted in gray color. Lower values are better}
    \label{fig:5}
\end{figure}

Software agreement for each single feature is measured by relative difference between the calculated feature values and the benchmark values provided by the IBSI (Eq. \ref{eq:5}). Significant differences in overall features are noted from CaPTk analysis (Fig. \ref{fig:5}), possibly from the effect of intensity shifting in the step of preprocessing. Good agreement is found between the other 6 software programs.

\begin{equation}
\label{eq:5}
    relative~difference = \frac{\mid feature~value - benchmark~value \mid}{benchmark~value}
\end{equation}

\begin{figure}
    \centering
    \includegraphics[width=129mm]{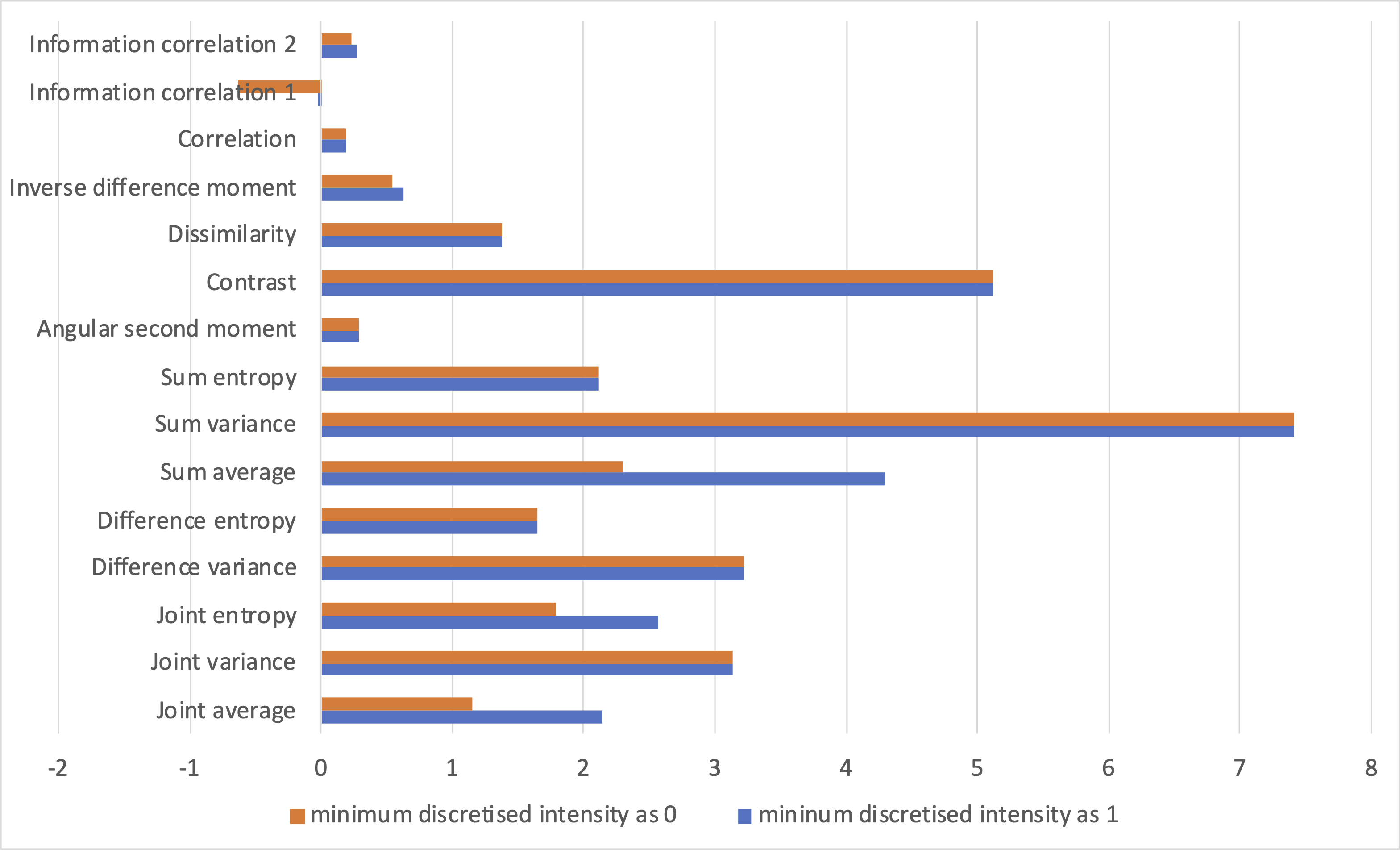}
    \caption{GLCM features calculated by A1 with two different settings of minimum intensities}
    \label{fig:6}
\end{figure}

To study the effect of intensity shifting, we modified the code for intensity shifting in our in house pipeline and produced the results of GLCM as an example. Results based on the intensity shifting in our in house pipeline and producing the results of GLCM as an example, revealed that  that some features (e.g. joint average, joint entropy, sum average and inverse difference moment) have significant deviation after adding the intensity shifting, while others (e.g. contrast, sum entropy) are unchanged (Fig. \ref{fig:6}).

\begin{figure}
    \includegraphics[width=129mm]{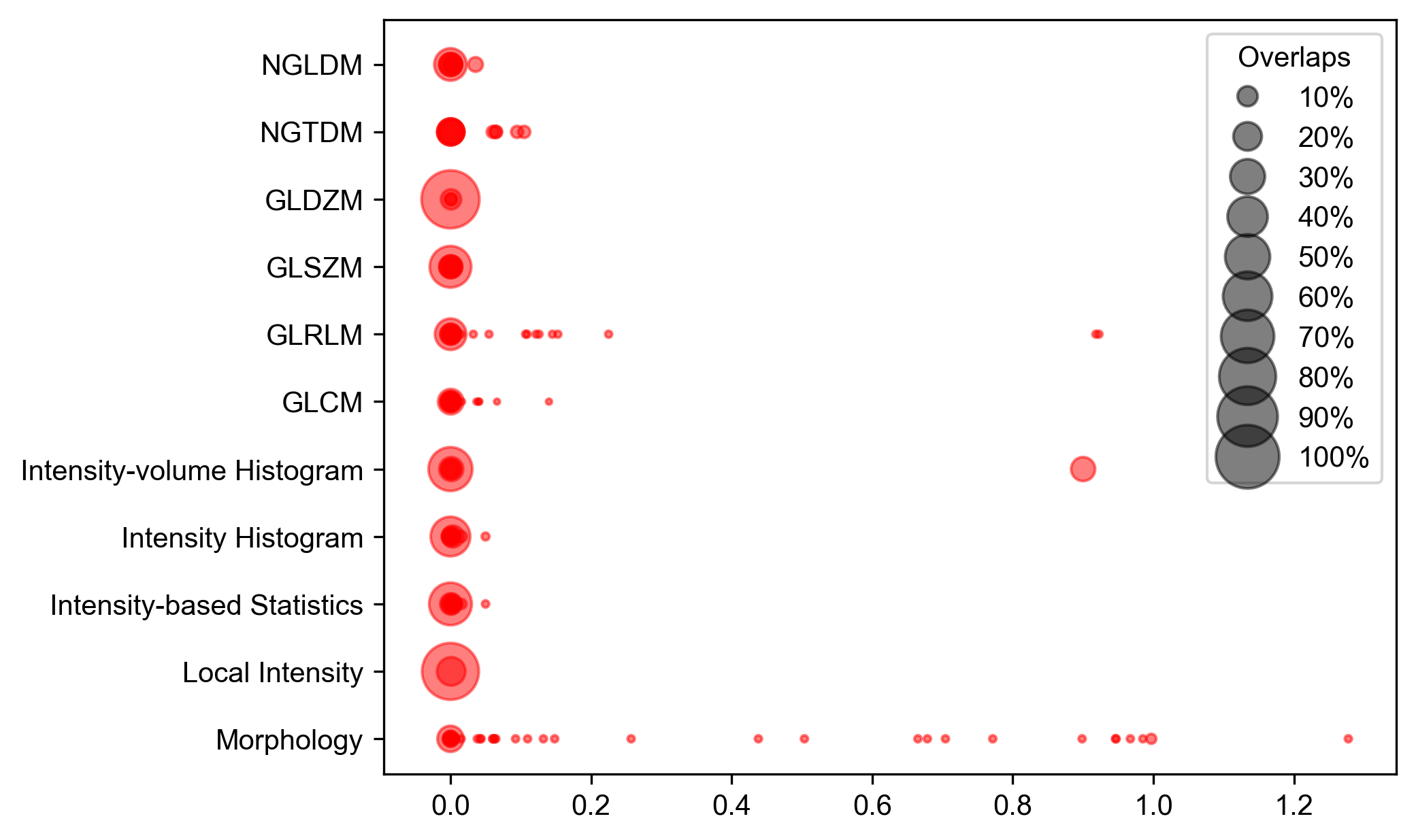}
    \caption{Scatter plot for the relative differences, concluding the rest 6 software programs (A1, Pyradiomics, MITK, LIFEX, SERA, A2). Larger size of points indicates that more points are overlapped at the underlying location (e.g. a point with Size 50\% means that 50\% of points within that feature class are equal). Large points around the location of 0 indicate good agreement across radiomics software programs}
    \label{fig:7}
\end{figure}

A scatter plot was used to analyze the agreement (measured by relative difference) among different feature classes (Fig. \ref{fig:7}). Since CaPTk preprocessed the intensity by shifting -1 unit, which is different to all other software, we considered all the other software programs in Fig. \ref{fig:7}. Good agreement is also observed among non-morphology features, while poor agreement is observed for morphology features

\begin{figure}
    \centering
    \subfloat[]{\includegraphics[width=84mm]{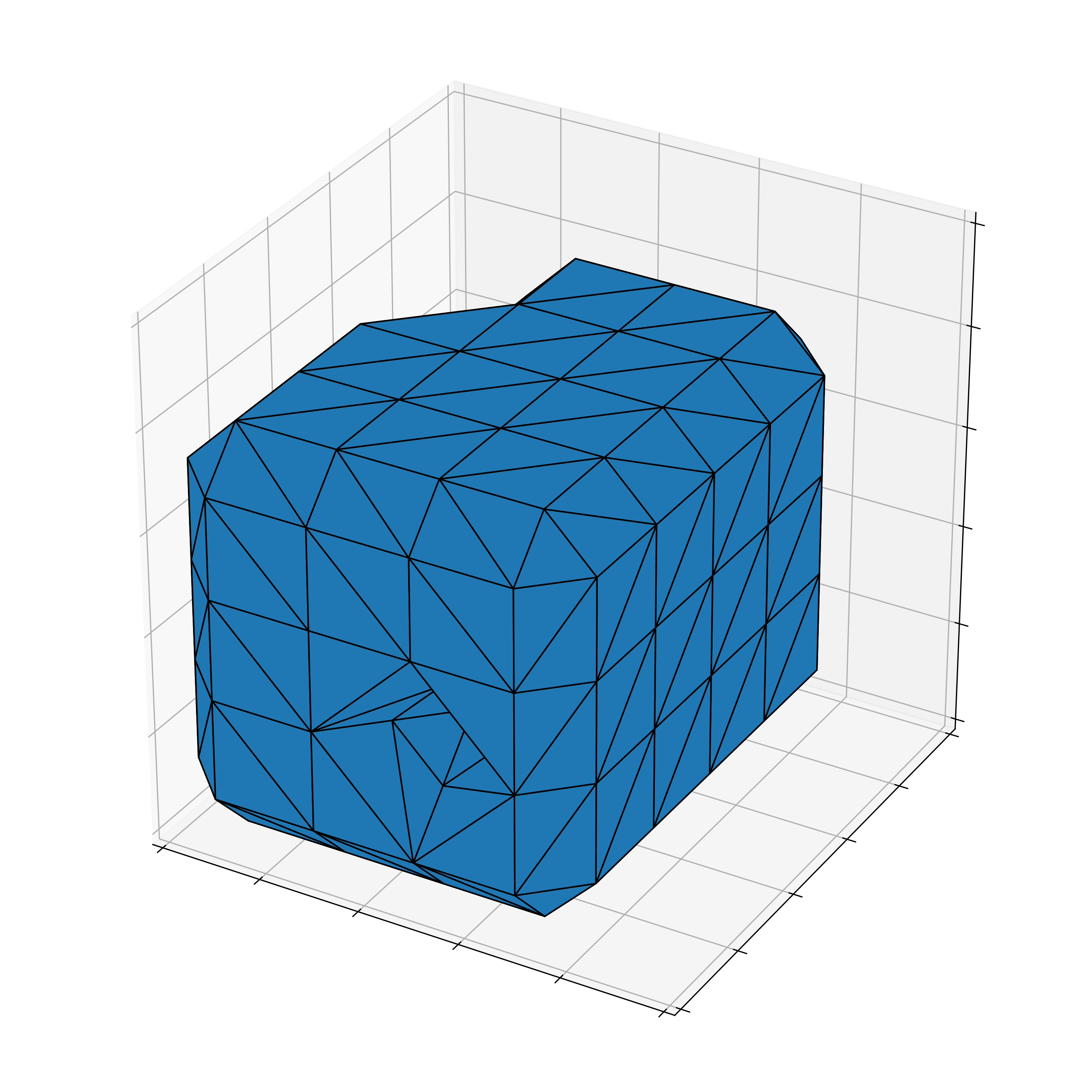}\label{fig:8-1}}
    \hfil
    \subfloat[]{\includegraphics[width=84mm]{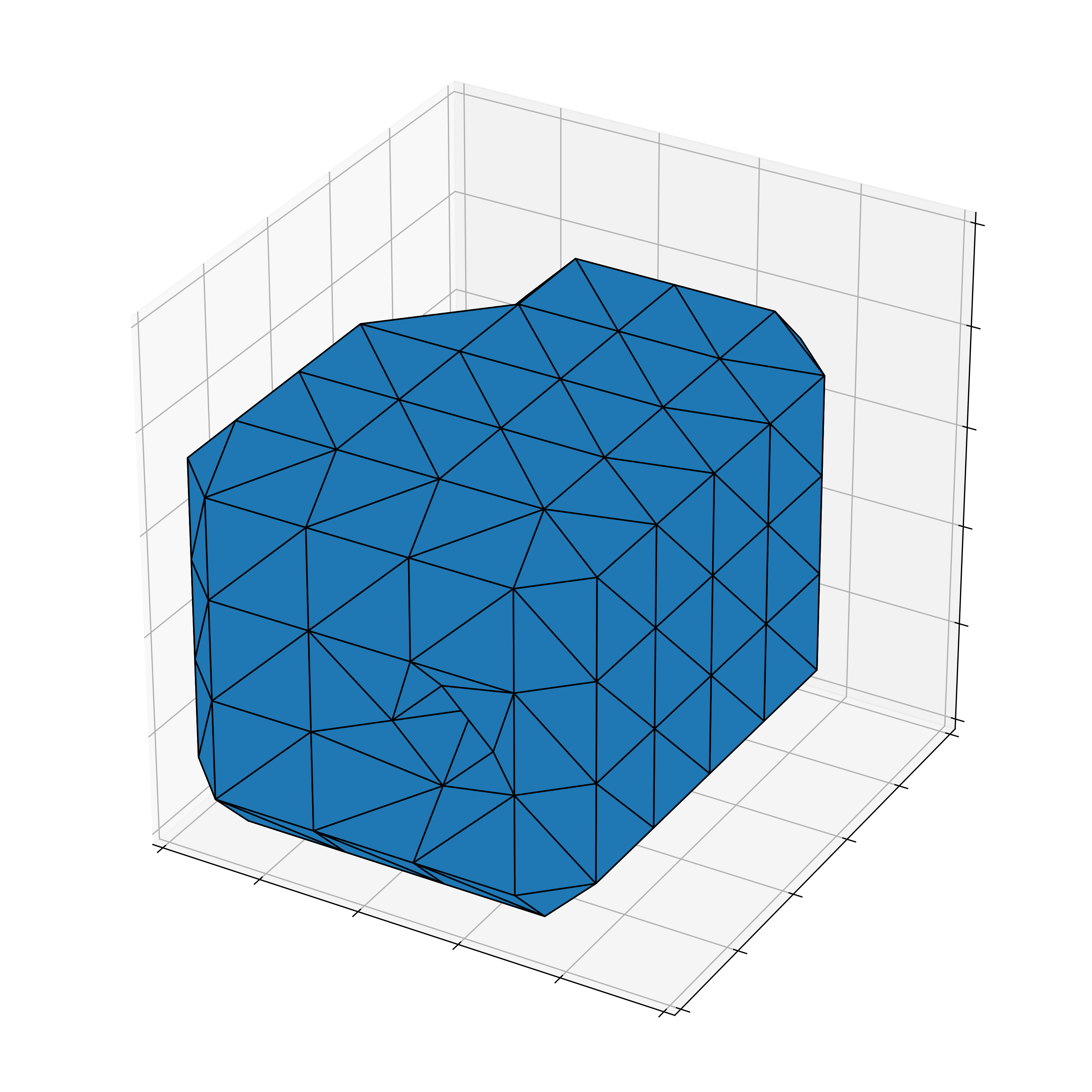}\label{fig:8-2}}
    \caption{(a) Meshing representations of the digital phantom generated by SERA. Value of volume is 556. (b) Meshing representations of the digital phantom generated by A2. Value of volume is 554}
    \label{fig:8}
\end{figure}

As morphology features are computed under the meshing representation, the computed meshes may also affect feature values. Among 7 software programs, we extracted the computed meshes from SERA and A2 at their intermediate steps 
(Fig. \ref{fig:8}). Other software programs are not able to give the meshes data, either because they are programmed as “end-to-end” pipelines or are not open-source. The meshing of SERA slightly differs from the one of A2. Values of volumes of them are also different (556 for SERA, 554 for A2).

\section{Discussion}
This study has demonstrated that the design and implementation of radiomics feature calculation vary across multiple software programs, using a single IBSI standard phantom. While in other early studies \cite{Monica2018,Bianchi2019,Foy2018}, the comparison and analyses of radiomics reproducibility, software agreement were reported using statistical hypothesis testing \cite{Monica2018,Foy2018}, spearman correlation coefficients \cite{Bianchi2019} or intraclass-correlation coefficients \cite{Foy2018} (ICC), using a set of features from datasets.

Different strategies of intensity discretization have an influence on the resulted feature values. As shown in Fig. \ref{fig:6}, two strategies (minimum intensity as 0 vs. minimum intensity as 1) produce different feature values. The reason for different values is that 0 is introduced in the computation, thus partial terms will be lost. Given the two different strategies, as an example, GLCM Joint Entropy will have two different mathematical definitions:

\begin{equation}
    - \sum _{i=0}^{N_g-1} \sum _{j=0}^{N_g-1} P(i,j) \log_2 P(i,j)
\end{equation}
\begin{equation}
    - \sum _{i=1}^{N_g} \sum _{j=1}^{N_g} P(i,j) \log_2 P(i,j)
\end{equation}
where $P(i,j)$ denotes the element of GLCM at the location of $i^{th}$ row and $j^{th}$ column. According to the definition of GLCM, both cases will result in identical $P(i,j)$, but the second-order statistical features may be affected because zeros are introduced in the calculation. Partial terms of co-occurrence will be lost in the multiplication operation with 0, causing relatively smaller feature values. This discretization strategy is not unique to just CaPTk; other studies also use the same mathematical definition, setting the minimum intensity to zero \cite{Mohanaiah2013ImageTF,Humeau2019,Gade2018}. By definition, the two discretization approaches given by the IBSI (Eq. \ref{eq:3} and Eq. \ref{eq:4}), add a constant unit of one to the minimum intensity avoid zeros encountered in the quantized image. Discretization should therefore adhere to the IBSI standard not only for reproducibility, but also for retaining the partial terms in the computation of the co-occurrence matrix to fully and precisely characterize the properties of ROI.

Disagreement among feature values also occurred frequently in the group of morphological features. By definition, the computation of morphology features is more complex than non-morphology features. Statistical and textural features are defined unequivocally on the 3D volume of ROI; hence the computation goes directly using matrix algebra. The computation of morphology features, by contrast, includes two stages. The source of feature value variation may come from the first stage, where volume data is transformed as meshes data. There are various algorithms for meshing approximation, where different algorithms may result in different meshes and therefore produce different feature values. As Fig. \ref{fig:8} shown, SERA and A2 may use different algorithms and give two different approximations and values of volumes. Therefore, careful consideration needs to be given to the choice of morphometric features used particularly in multi-center radiomics analysis.

The “popularity” of each features is important for analysis of software agreement. With more software giving a same feature, the analysis of it is more reliable. Within 173 IBSI-standardized features, Fig. 2 shows that the 6 publicly available software programs share 1/29 morphology feature, 7/50 statistics and histogram features and 21/94 texture features. Furthermore, both ad hoc metrics, $P_1$ and $P_2$ , have lower values for morphology features (Fig. \ref{fig:3}). These indicate that relatively less attention is provided to morphology properties of an ROI compared to its non-morphological properties. This may be true considering the  computational complexity behind extracting morphology features, i.e., the algorithms and programming of meshing representations are usually challenging, tedious and error prone \cite{Lorensen1987,Chernikov2014,DELIBASIS2001343,RAJON2003411}. Taken together these reasons justify the poor representation and feature inconsistency of morphological metrics within radiomics software.

The primary aim of most radiomics studies is to identify significant features and build discriminant models. Therefore, it is desirable to improve the model performance by including more candidate features in the statistical analysis. With more features, in general, it is more possible to find a better subset of radiomics features to improve the performance, achieving higher accuracy, area under the curve to aid discrimination. In addition, morphology features show significant prognostic power in the application of various cancer types \cite{Parmar2015,CUOCOLO2019144,Zhu2015}. Therefore, despite the computational complexity in extracting morphology features which are rich in information, possibly distinct from that provided by non-morphological metrics, they should be included in analysis. The results reveal that current software programs lack standardization. Benchmarking may be important for a reproducible radiomics study. The IBSI digital phantom is a good resource for validating programs with the provided benchmark values for each proposed feature. In most studies, images are analyzed with preprocessing (normalization, transformation, discretization). While the phantom is not be able to contribute to the benchmarking of preprocessing directly because it does not require additional processing, it may be helpful indirectly by checking the basic features in Table \ref{tab:6}, where the difference of discretization approach are found. The intensity-based statistical features are calculated right after interpolation, which can be used for benchmarking of image interpolation; intensity histogram features are calculated right after gray level discretization, which can be used for benchmarking of both interpolation and discretization  (Fig. \ref{fig:4-1}).

We conclude that future radiomic studies should provide implementation details about the radiomics software such as the name of the software, its version, whether it is open-source or custom-built, the programming language used to develop it etc., so that appropriate comparisons of the results across different radiomics studies can be performed. In addition, consistent terminology of the features should be used, ideally adhering to the IBSI terminologies. Features not used in the IBSI list should be thoroughly defined and supported by references, so future studies can validate them.

\section{Limitation}
This study evaluated only 6 publicly available and 1 in-house program. Adding more software programs may strengthen our findings. For some programs, computation and code running tracking may not available, leading to the difficulty in matching the program available to the program used to publish results. While some programs are readable (e.g. MATLAB toolbox), other programs only provide graphical user interface (GUI) for the convenience of non-professional users and codes are not visible. As an example, LIFEx is programmed in Java, and the source and reason of features deviation is not easy to understand.

Further analysis could focus on the feature comparison on real-world datasets. With a larger range of gray levels in the real cases, different imaging modalities, the features agreement may be different. Most studies employ image enhancement, augmentation, and transformation before feature extraction to find more significant features, such as image filtering, edge enhancement. These operations may also affect the degree of variation across radiomics programs. 

\section{Conclusion}
We evaluated feature agreement in the use of 7 different radiomics software programs. While most first-order and second-order texture features show satisfactory agreement, morphological features show significant variation. In addition, most programs are relatively weak on morphology analysis possibly owing to its computation complexity. Further work is necessary to standardize the calculation of radiomics features across different radiomics software, which is one of the first steps towards the clinical translation of radiomic analysis.

\begin{table}
    \centering
    \caption{Three major categories of 11 classes of IBSI-standardized features. Morphology features stand alone as a single category.}
    \label{tab:1}
    \begin{tabular}{lll}
        \hline\noalign{\smallskip}
        Statistical/histogram features & Texture features & Morphology (shape) features  \\
        \noalign{\smallskip}\hline\noalign{\smallskip}
        Local intensity & GLCM &  Morphology features\\
        Intensity-based statistic & GLRLM &  \\
        Intensity histogram & GLSZM &  \\
        Intensity-volume histogram & GLDZM &  \\
         & NGTDM &  \\
         & NGLDM &  \\
        \noalign{\smallskip}\hline
    \end{tabular}
\end{table}

\begin{landscape}

\begin{table}
\caption{Radiomics features}
\label{tab:2}       
\begin{tabular}{lllllllll}
\hline\noalign{\smallskip}
& IBSI-standardized & A1 & Pyradiomics & MITK & LIFEx & SERA & CaPTk & A2\\
\noalign{\smallskip}\hline\noalign{\smallskip}
Morphological features & 29 & 1 & 14 & 20 & 2 & 25 & 3 & 27 \\
Local intensity features & 2 & 0 & 0 & 2 & 0 & 2 & 0 & 1   \\
Intensity-based statistical features & 18 & 6 & 15 & 18 & 5 & 18 & 15 & 18  \\
Intensity histogram features & 23 & 0 & 2 & 21 & 2 & 19 & 17 & 23  \\
Intensity-volume histogram features & 7 & 0 & 0 & 7 & 0 & 7 & 0 & 7  \\
GLCM & 25 & 15 & 23 & 25 & 6 & 25 & 6 & 25  \\
GLRLM & 16	& 11 & 16 & 16 & 11 & 16 & 4 & 16  \\
GLSZM  & 16 & 16 & 16 & 16 & 11 & 16 & 16 & 16  \\
GLDZM & 16 & 0 & 0 & 0 & 0 & 16 & 0 & 16 \\
NGTDM & 5 & 5 & 5 & 5 & 3 & 5 & 5 & 5 \\
NGLDM & 16 & 16 & 13 & 16 & 0 & 16 & 16 & 16 \\

\noalign{\smallskip}\hline
\end{tabular}
\end{table}

\begin{table}
\caption{Radiomics software programs}
\label{tab:3}       
\begin{tabular}{llll}
\hline\noalign{\smallskip}
& Language & Supported Operating System & Significant Dependent Library\\
\noalign{\smallskip}\hline\noalign{\smallskip}
A1 & MATLAB & Windows, Linux, MacOS & None\\
Pyradiomics & Python & Windows, Linux, MacOS & SimpleITK\\
MITK & C++ & Windows, Linux, MacOS & the Insight Toolkit (ITK), the Visualization Toolkit (VTK)\\
LIFEx & Java & Windows, Linux, MacOS & None\\
SERA & MATLAB & Windows, Linux, MacOS & None\\
CaPTk & C++ & Windows, Linux, MacOS & OpenCV, ITK, VTK\\
A2 & MATLAB & Windows, Linux, MacOS & None\\

\noalign{\smallskip}\hline
\end{tabular}
\end{table}

\begin{table}
\small
\caption{Algorithm references across different software}
\label{tab:4}       
\begin{tabular}{lllllllll}
\hline\noalign{\smallskip}
& A1 & Pyradiomics & MITK & LIFEx & SERA & CaPTk & A2 & IBSI recommendation\\
\noalign{\smallskip}\hline\noalign{\smallskip}
Morphological features & - & Lorensen\cite{Lorensen1987} &- &- & IBSI\cite{Zwanenburg2020} & Valli{\`e}res\cite{Valli_res_2015} & IBSI\cite{Zwanenburg2020} & Lorensen\cite{Lorensen1987} \\
 & & & & & Varn{\"a}s\cite{Varnas2004} & Thibault\cite{THIBAULT2013} & & Lewiner\cite{Lewiner2003}\\[5pt]
Local intensity features & - & - & - & - & - & - & IBSI\cite{Zwanenburg2020} & Wahl\cite{Wahl2009}   \\
 &  &  &  &  &  &  &  & Frings\cite{Frings2014}   \\[5pt]
Intensity-based statistical features & - & - & - & -& IBSI\cite{Zwanenburg2020} & - & IBSI\cite{Zwanenburg2020} & -  \\[5pt]
Intensity histogram features & - & - & - & - & - & Macyszyn\cite{Macyszyn2015} & IBSI\cite{Zwanenburg2020} & - \\[5pt]
Intensity-volume histogram features & - & - & - & - & IBSI\cite{Zwanenburg2020} & - & IBSI\cite{Zwanenburg2020} & El Naqa\cite{ELNAQA20091162}\\[5pt]
GLCM & Haralick\cite{Haralick1973} & Haralick\cite{Haralick1973} & - & Haralick\cite{Haralick1973} & Haralick\cite{Haralick1973} & Haralick \cite{Haralick1973} & IBSI\cite{Zwanenburg2020} & Haralick\cite{Haralick1973}  \\
 &  &  &  &  & IBSI\cite{Zwanenburg2020} &  &  & Unser\cite{Unser1986}  \\[5pt]
GLRLM & Galloway\cite{GALLOWAY1975172}	& Galloway\cite{GALLOWAY1975172} & - & Xu\cite{Xu2004} & Galloway\cite{GALLOWAY1975172} & Galloway\cite{GALLOWAY1975172} & IBSI\cite{Zwanenburg2020} & Galloway\cite{GALLOWAY1975172}  \\
 & & Chu\cite{CHU1990415} &  &  &  &  &  &   \\
 & & Xu\cite{Xu2004} &  &  &  &  &  &   \\
 & & Tang\cite{Tang1998} &  &  &  &  &  &   \\
 & & Tustison\cite{Tustison2009} &  &  &  &  &  &   \\[5pt]
GLSZM  & Thibault\cite{Thibault2009} & Thibault\cite{Thibault2009} & - & Thibault\cite{Thibault2009} & IBSI\cite{Zwanenburg2020} & Galloway\cite{GALLOWAY1975172} & IBSI\cite{Zwanenburg2020} & Thibault\cite{Thibault2014}  \\
 &  &  &  &  &  & Tang\cite{Tang1998} &  &   \\[5pt]
GLDZM & - & - & - & - & IBSI\cite{Zwanenburg2020} & - & IBSI\cite{Zwanenburg2020} & Thibault\cite{Thibault2014} \\[5pt]
NGTDM & Amadasun\cite{Amadasun1989} & Amadasun\cite{Amadasun1989} & - & Amadasun\cite{Amadasun1989} & IBSI\cite{Zwanenburg2020} & Amadasun\cite{Amadasun1989} & IBSI\cite{Zwanenburg2020} & Amadasun\cite{Amadasun1989} \\[5pt]
NGLDM & Sun\cite{SUN1983341} & Sun\cite{SUN1983341} & - & - & IBSI\cite{Zwanenburg2020} & - & IBSI\cite{Zwanenburg2020} & Sun\cite{SUN1983341} \\

\noalign{\smallskip}\hline
\end{tabular}
\end{table}

\begin{table}
\caption{Computation parameters}
\label{tab:5}       
\begin{tabular}{lllllllll}
\hline\noalign{\smallskip}
 & & A1 & Pyradiomics & MITK & LIFEx & SERA & CaPTk & A2   \\
\noalign{\smallskip}\hline\noalign{\smallskip}
Interpolation & & - & - & - & 2.0 $\times$ 2.0 $\times$ 2.0 mm & - & 2.0 $\times$ 2.0 $\times$ 2.0 mm & - \\
Discretization & & - & - & - & Fix bin size of 1.0 & - & Fix bin number of 6 & - \\
GLCM & Neighborhood Dimension & 3D & 3D & 3D & 3D & 3D & 3D  & 3D \\
 & Distance & 1 & 1 & 1 & 1 & 1 & 1 & 1 \\
 & Aggregation Method & Merging & Merging & Merging & Averaing & Merging & Merging & Merging \\
GLRLM & Neighborhood Dimension & 3D & 3D & 3D & 3D & 3D & 3D  & 3D \\
 & Distance & 1 & 1 & 1 & 1 & 1 & 1 & 1 \\
 & Aggregation Method & Merging & Merging & Merging & Averaing & Merging & Merging & Merging \\
GLSZM & Neighborhood Dimension & 3D & 3D & 3D & 3D & 3D & 3D  & 3D \\
 & Distance & 1 & 1 & 1 & 1 & 1 & 1 & 1 \\
GLDZM & Neighborhood Dimension & - & - & - & - & 3D & -  & 3D \\ 
 & Distance & - & - & - & - & 1 & - & 1 \\
NGTDM & Neighborhood Dimension & 3D & 3D & 3D & 3D & 3D & 3D  & 3D \\
 & Distance & 1 & 1 & 1 & 1 & 1 & 1 & 1 \\
NGLDM & Neighborhood Dimension & 3D & 3D & 3D & - & 3D & 3D  & 3D \\ 
 & Distance & 1 & 1 & 1 & - & 1 & 1 & 1 \\
 & Coarseness & 0 & 0 & 0 & - & 0 & 0 & 0 \\ 

\noalign{\smallskip}\hline
\end{tabular}
\end{table}

\begin{table}
\caption{Basic features}
\label{tab:6}       
\begin{tabular}{lllllllll}
\hline\noalign{\smallskip}
 & A1 & Pyradiomics & MITK & LIFEx & SERA & CaPTk & A2 & IBSI Benchmark Value  \\
\noalign{\smallskip}\hline\noalign{\smallskip}
Mean discretised intensity & - & - & 2.1643 & - & 2.1486 & 1.1486 & 2.1486 & 2.15\\
Median discretised intensity & - & - & 1.0156 & - & 1 & 0 & 1 & 1\\
Minimum discretised intensity & - & - & 1 & - & 1 & 0 & 1 & 1\\
Maximum discretised intensity & - & - & 6 & - & 6 & 5 & 6 & 6\\
Discretised intensity range & - & - & 5 & - & 5 & 5 & 5 & 5\\

\noalign{\smallskip}\hline
\end{tabular}
\end{table}

\end{landscape}

\bibliographystyle{spmpsci}
\bibliography{manuscript.bib}{}

\begin{thebibliography}{10}
\providecommand{\url}[1]{{#1}}
\providecommand{\urlprefix}{URL }
\expandafter\ifx\csname urlstyle\endcsname\relax
  \providecommand{\doi}[1]{DOI~\discretionary{}{}{}#1}\else
  \providecommand{\doi}{DOI~\discretionary{}{}{}\begingroup
  \urlstyle{rm}\Url}\fi

\bibitem{Court2016}
Court, L.E., Fave, X., Mackin, D., Lee, J., Yang, J., Zhang, L.: Computational
  resources for radiomics.
\newblock Translational Cancer Research \textbf{5}(4) (2016).
\newblock \urlprefix\url{http://tcr.amegroups.com/article/view/8409}

\bibitem{Lee2019}
Lee, S.H., Cho, H.h., Lee, H.Y., Park, H.: Clinical impact of variability on ct
  radiomics and suggestions for suitable feature selection: a focus on lung
  cancer.
\newblock Cancer Imaging \textbf{19}(1), 54 (2019).
\newblock \doi{10.1186/s40644-019-0239-z}.
\newblock \urlprefix\url{https://doi.org/10.1186/s40644-019-0239-z}

\bibitem{Monica2018}
B{\'e}resov{\'a}, M., Forg{\'a}cs, A., Bujdos{\'o}, B., Sz{\'e}kely, A., Varga,
  J., Ber{\'e}nyi, E., Balkay, L.: Comparing the reliability of biomedical
  texture analysis tools on different image types.
\newblock Acta Polytechnica Hungarica \textbf{15}(7), 29--48 (2018).
\newblock \doi{10.12700/APH.15.7.2018.7.2}

\bibitem{Bianchi2019}
Bianchi, J., Gonçalves, J.R., Ruellas, A.C.d.O., Vimort, J.B., Yatabe, M.,
  Paniagua, B., Hernandez, P., Benavides, E., Soki, F.N., Cevidanes, L.H.S.:
  Software comparison to analyze bone radiomics from high resolution cbct scans
  of mandibular condyles.
\newblock Dentomaxillofacial Radiology \textbf{48}(6), 20190049 (2019).
\newblock \doi{10.1259/dmfr.20190049}.
\newblock \urlprefix\url{https://doi.org/10.1259/dmfr.20190049}.
\newblock PMID: 31075043

\bibitem{Foy2018}
Foy, J.J., Robinson, K.R., Li, H., Giger, M.L., Al-Hallaq, H., Armato, S.G.:
  {Variation in algorithm implementation across radiomics software}.
\newblock Journal of Medical Imaging \textbf{5}(4), 1 -- 10 (2018).
\newblock \doi{10.1117/1.JMI.5.4.044505}.
\newblock \urlprefix\url{https://doi.org/10.1117/1.JMI.5.4.044505}

\bibitem{Shafig2017}
Shafiq-ul Hassan, M., Zhang, G.G., Latifi, K., Ullah, G., Hunt, D.C.,
  Balagurunathan, Y., Abdalah, M.A., Schabath, M.B., Goldgof, D.G., Mackin, D.,
  Court, L.E., Gillies, R.J., Moros, E.G.: Intrinsic dependencies of ct
  radiomic features on voxel size and number of gray levels.
\newblock Medical Physics \textbf{44}(3), 1050--1062 (2017).
\newblock \doi{10.1002/mp.12123}.
\newblock
  \urlprefix\url{https://aapm.onlinelibrary.wiley.com/doi/abs/10.1002/mp.12123}

\bibitem{Bogowicz2016}
Bogowicz, M., Riesterer, O., Bundschuh, R.A., Veit-Haibach, P., Hüllner, M.,
  Studer, G., Stieb, S., Glatz, S., Pruschy, M., Guckenberger, M.,
  Tanadini-Lang, S.: Stability of radiomic features in {CT} perfusion maps.
\newblock Physics in Medicine and Biology \textbf{61}(24), 8736--8749 (2016).
\newblock \doi{10.1088/1361-6560/61/24/8736}.
\newblock
  \urlprefix\url{https://doi.org/10.1088\%2F1361-6560\%2F61\%2F24\%2F8736}

\bibitem{Kumar2012}
Kumar, V., Gu, Y., Basu, S., Berglund, A., Eschrich, S.A., Schabath, M.B.,
  Forster, K., Aerts, H.J., Dekker, A., Fenstermacher, D., Goldgof, D.B., Hall,
  L.O., Lambin, P., Balagurunathan, Y., Gatenby, R.A., Gillies, R.J.:
  Radiomics: the process and the challenges.
\newblock Magnetic Resonance Imaging \textbf{30}(9), 1234 -- 1248 (2012).
\newblock \doi{https://doi.org/10.1016/j.mri.2012.06.010}.
\newblock
  \urlprefix\url{http://www.sciencedirect.com/science/article/pii/S0730725X12002202}.
\newblock Quantitative Imaging in Cancer

\bibitem{Schwier2019}
Schwier, M., van Griethuysen, J., Vangel, M.G., Pieper, S., Peled, S., Tempany,
  C., Aerts, H.J.W.L., Kikinis, R., Fennessy, F.M., Fedorov, A.: Repeatability
  of multiparametric prostate mri radiomics features.
\newblock Scientific Reports \textbf{9}(1), 9441 (2019).
\newblock \doi{10.1038/s41598-019-45766-z}.
\newblock \urlprefix\url{https://doi.org/10.1038/s41598-019-45766-z}

\bibitem{Mackin2015}
Mackin, D., Fave, X., Zhang, L., Fried, D., Yang, J., Taylor, B.,
  Rodriguez-Rivera, E., Dodge, C., Jones, A.K., Court, L.: Measuring computed
  tomography scanner variability of radiomics features.
\newblock Investigative radiology \textbf{50}(11), 757--765 (2015).
\newblock \doi{10.1097/RLI.0000000000000180}.
\newblock \urlprefix\url{https://doi.org/10.1097/RLI.0000000000000180}

\bibitem{vanVelden2016}
van Velden, F.H.P., Kramer, G.M., Frings, V., Nissen, I.A., Mulder, E.R.,
  de~Langen, A.J., Hoekstra, O.S., Smit, E.F., Boellaard, R.: Repeatability of
  radiomic features in non-small-cell lung cancer [18f]fdg-pet/ct studies:
  Impact of reconstruction and delineation.
\newblock Molecular Imaging and Biology \textbf{18}(5), 788--795 (2016).
\newblock \doi{10.1007/s11307-016-0940-2}.
\newblock \urlprefix\url{https://doi.org/10.1007/s11307-016-0940-2}

\bibitem{Zwanenburg2020}
Zwanenburg, A., Vallières, M., Abdalah, M.A., Aerts, H.J.W.L., Andrearczyk,
  V., Apte, A., Ashrafinia, S., Bakas, S., Beukinga, R.J., Boellaard, R.,
  et~al.: The image biomarker standardization initiative: Standardized
  quantitative radiomics for high-throughput image-based phenotyping.
\newblock Radiology \textbf{295}(2), 328–338 (2020).
\newblock \doi{10.1148/radiol.2020191145}.
\newblock \urlprefix\url{http://dx.doi.org/10.1148/radiol.2020191145}

\bibitem{van_Griethuysene2017}
van Griethuysen, J.J., Fedorov, A., Parmar, C., Hosny, A., Aucoin, N., Narayan,
  V., Beets-Tan, R.G., Fillion-Robin, J.C., Pieper, S., Aerts, H.J.:
  Computational radiomics system to decode the radiographic phenotype.
\newblock Cancer Research \textbf{77}(21), e104--e107 (2017).
\newblock \doi{10.1158/0008-5472.CAN-17-0339}.
\newblock \urlprefix\url{https://cancerres.aacrjournals.org/content/77/21/e104}

\bibitem{Echegaray2018}
Echegaray, S., Bakr, S., Rubin, D.L., Napel, S.: Quantitative image feature
  engine (qife): an open-source, modular engine for 3d quantitative feature
  extraction from volumetric medical images.
\newblock Journal of Digital Imaging \textbf{31}(4), 403--414 (2018).
\newblock \doi{10.1007/s10278-017-0019-x}.
\newblock \urlprefix\url{https://doi.org/10.1007/s10278-017-0019-x}

\bibitem{Aerts2014}
Aerts, H.J.W.L., Velazquez, E.R., Leijenaar, R.T.H., Parmar, C., Grossmann, P.,
  Carvalho, S., Bussink, J., Monshouwer, R., Haibe-Kains, B., Rietveld, D.,
  Hoebers, F., Rietbergen, M.M., Leemans, C.R., Dekker, A., Quackenbush, J.,
  Gillies, R.J., Lambin, P.: Decoding tumour phenotype by noninvasive imaging
  using a quantitative radiomics approach.
\newblock Nature Communications \textbf{5}(1), 4006 (2014).
\newblock \doi{10.1038/ncomms5006}.
\newblock \urlprefix\url{https://doi.org/10.1038/ncomms5006}

\bibitem{Wu2016}
Wu, W., Parmar, C., Grossmann, P., Quackenbush, J., Lambin, P., Bussink, J.,
  Mak, R., Aerts, H.J.W.L.: Exploratory study to identify radiomics classifiers
  for lung cancer histology.
\newblock Frontiers in Oncology \textbf{6}, 71 (2016).
\newblock \doi{10.3389/fonc.2016.00071}.
\newblock
  \urlprefix\url{https://www.frontiersin.org/article/10.3389/fonc.2016.00071}

\bibitem{COROLLER2015345}
Coroller, T.P., Grossmann, P., Hou, Y., Velazquez], E.R., Leijenaar, R.T.,
  Hermann, G., Lambin, P., Haibe-Kains, B., Mak, R.H., Aerts, H.J.: Ct-based
  radiomic signature predicts distant metastasis in lung adenocarcinoma.
\newblock Radiotherapy and Oncology \textbf{114}(3), 345 -- 350 (2015).
\newblock \doi{https://doi.org/10.1016/j.radonc.2015.02.015}.
\newblock
  \urlprefix\url{http://www.sciencedirect.com/science/article/pii/S0167814015001073}

\bibitem{Dou2018}
Dou, T.H., Coroller, T.P., van Griethuysen, J.J.M., Mak, R.H., Aerts, H.J.W.L.:
  Peritumoral radiomics features predict distant metastasis in locally advanced
  nsclc.
\newblock PloS one \textbf{13}(11), e0206108--e0206108 (2018).
\newblock \doi{10.1371/journal.pone.0206108}.
\newblock \urlprefix\url{https://doi.org/10.1371/journal.pone.0206108}

\bibitem{vanGriethuysen2020}
van Griethuysen, J.J.M., Lambregts, D.M.J., Trebeschi, S., Lahaye, M.J.,
  Bakers, F.C.H., Vliegen, R.F.A., Beets, G.L., Aerts, H.J.W.L., Beets-Tan,
  R.G.H.: Radiomics performs comparable to morphologic assessment by expert
  radiologists for prediction of response to neoadjuvant chemoradiotherapy on
  baseline staging mri in rectal cancer.
\newblock Abdominal Radiology \textbf{45}(3), 632--643 (2020).
\newblock \doi{10.1007/s00261-019-02321-8}.
\newblock \urlprefix\url{https://doi.org/10.1007/s00261-019-02321-8}

\bibitem{GOTZ2019108}
Götz, M., Nolden, M., Maier-Hein, K.: Mitk phenotyping: An open-source
  toolchain for image-based personalized medicine with radiomics.
\newblock Radiotherapy and Oncology \textbf{131}, 108 -- 111 (2019).
\newblock \doi{https://doi.org/10.1016/j.radonc.2018.11.021}.
\newblock
  \urlprefix\url{http://www.sciencedirect.com/science/article/pii/S0167814018336156}

\bibitem{Kickingereder5765}
Kickingereder, P., G{\"o}tz, M., Muschelli, J., Wick, A., Neuberger, U.,
  Shinohara, R.T., Sill, M., Nowosielski, M., Schlemmer, H.P., Radbruch, A.,
  Wick, W., Bendszus, M., Maier-Hein, K.H., Bonekamp, D.: Large-scale radiomic
  profiling of recurrent glioblastoma identifies an imaging predictor for
  stratifying anti-angiogenic treatment response.
\newblock Clinical Cancer Research \textbf{22}(23), 5765--5771 (2016).
\newblock \doi{10.1158/1078-0432.CCR-16-0702}.
\newblock
  \urlprefix\url{https://clincancerres.aacrjournals.org/content/22/23/5765}

\bibitem{Kickingereder2016}
Kickingereder, P., Burth, S., Wick, A., Götz, M., Eidel, O., Schlemmer, H.P.,
  Maier-Hein, K.H., Wick, W., Bendszus, M., Radbruch, A., Bonekamp, D.:
  Radiomic profiling of glioblastoma: Identifying an imaging predictor of
  patient survival with improved performance over established clinical and
  radiologic risk models.
\newblock Radiology \textbf{280}(3), 880--889 (2016).
\newblock \doi{10.1148/radiol.2016160845}.
\newblock \urlprefix\url{https://doi.org/10.1148/radiol.2016160845}.
\newblock PMID: 27326665

\bibitem{Nioche}
Nioche, C., Orlhac, F., Boughdad, S., Reuz{\'e}, S., Goya-Outi, J., Robert, C.,
  Pellot-Barakat, C., Soussan, M., Frouin, F., Buvat, I.: Lifex: A freeware for
  radiomic feature calculation in multimodality imaging to accelerate advances
  in the characterization of tumor heterogeneity.
\newblock Cancer Research \textbf{78}(16), 4786--4789 (2018).
\newblock \doi{10.1158/0008-5472.CAN-18-0125}.
\newblock \urlprefix\url{https://cancerres.aacrjournals.org/content/78/16/4786}

\bibitem{Nioche01052017}
Nioche, C., Orlhac, F., Boughdad, S., Reuze, S., Soussan, M., Robert, C.,
  Barakat, C., Buvat, I.: A freeware for tumor heterogeneity characterization
  in pet, spect, ct, mri and us to accelerate advances in radiomics.
\newblock Journal of Nuclear Medicine \textbf{58}(supplement 1), 1316 (2017)

\bibitem{Nardone2018}
Nardone, V., Tini, P., Nioche, C., Mazzei, M.A., Carfagno, T., Battaglia, G.,
  Pastina, P., Grassi, R., Sebaste, L., Pirtoli, L.: Texture analysis as a
  predictor of radiation-induced xerostomia in head and neck patients
  undergoing imrt.
\newblock La radiologia medica \textbf{123}(6), 415--423 (2018).
\newblock \doi{10.1007/s11547-017-0850-7}.
\newblock \urlprefix\url{https://doi.org/10.1007/s11547-017-0850-7}

\bibitem{Ashrafinia2019}
Ashrafinia, S.: Quantitative nuclear medicine imaging using advanced image
  reconstruction and radiomics.
\newblock Ph.D. thesis, John Hopskins University (2019)

\bibitem{Ashrafinia01052018}
Ashrafinia, S., Dalaie, P., Yan, R., Huang, P., Pomper, M., Schindler, T.,
  Rahmim, A.: Application of texture and radiomics analysis to clinical
  myocardial perfusion spect imaging.
\newblock Journal of Nuclear Medicine \textbf{59}(supplement 1), 94 (2018)

\bibitem{Du2020}
Du, D., Feng, H., Lv, W., Ashrafinia, S., Yuan, Q., Wang, Q., Yang, W., Feng,
  Q., Chen, W., Rahmim, A., Lu, L.: Machine learning methods for optimal
  radiomics-based differentiation between recurrence and inflammation:
  Application to nasopharyngeal carcinoma post-therapy pet/ct images.
\newblock Molecular Imaging and Biology \textbf{22}(3), 730--738 (2020).
\newblock \doi{10.1007/s11307-019-01411-9}.
\newblock \urlprefix\url{https://doi.org/10.1007/s11307-019-01411-9}

\bibitem{Davatzikos2018}
Davatzikos, C., Rathore, S., Bakas, S., Pati, S., Bergman, M., Kalarot, R.,
  Sridharan, P., Gastounioti, A., Jahani, N., Cohen, E., Akbari, H., Tunc, B.,
  Doshi, J., Parker, D., Hsieh, M., Sotiras, A., Li, H., Ou, Y., Doot, R.K.,
  Bilello, M., Fan, Y., Shinohara, R.T., Yushkevich, P., Verma, R., Kontos, D.:
  {Cancer imaging phenomics toolkit: quantitative imaging analytics for
  precision diagnostics and predictive modeling of clinical outcome}.
\newblock Journal of Medical Imaging \textbf{5}(1), 1 -- 21 (2018).
\newblock \doi{10.1117/1.JMI.5.1.011018}.
\newblock \urlprefix\url{https://doi.org/10.1117/1.JMI.5.1.011018}

\bibitem{Pati2020}
Pati, S., Singh, A., Rathore, S., Gastounioti, A., Bergman, M., Ngo, P., Ha,
  S.M., Bounias, D., Minock, J., Murphy, G., Li, H., Bhattarai, A., Wolf, A.,
  Sridaran, P., Kalarot, R., Akbari, H., Sotiras, A., Thakur, S.P., Verma, R.,
  Shinohara, R.T., Yushkevich, P., Fan, Y., Kontos, D., Davatzikos, C., Bakas,
  S.: The cancer imaging phenomics toolkit (captk): Technical overview.
\newblock In: A.~Crimi, S.~Bakas (eds.) Brainlesion: Glioma, Multiple
  Sclerosis, Stroke and Traumatic Brain Injuries, pp. 380--394. Springer
  International Publishing, Cham (2020)

\bibitem{Rathore2018}
Rathore, S., M.D., H.A., Doshi, J., M.D., G.S., Rozycki, M., M.D., M.B., M.D.,
  R.A.L., Davatzikos, C.A.: {Radiomic signature of infiltration in peritumoral
  edema predicts subsequent recurrence in glioblastoma: implications for
  personalized radiotherapy planning}.
\newblock Journal of Medical Imaging \textbf{5}(2), 1 -- 10 (2018).
\newblock \doi{10.1117/1.JMI.5.2.021219}.
\newblock \urlprefix\url{https://doi.org/10.1117/1.JMI.5.2.021219}

\bibitem{Valli_res_2015}
Valli{\`{e}}res, M., Freeman, C.R., Skamene, S.R., Naqa, I.E.: A radiomics
  model from joint {FDG}-{PET} and {MRI} texture features for the prediction of
  lung metastases in soft-tissue sarcomas of the extremities.
\newblock Physics in Medicine and Biology \textbf{60}(14), 5471--5496 (2015).
\newblock \doi{10.1088/0031-9155/60/14/5471}.
\newblock
  \urlprefix\url{https://doi.org/10.1088\%2F0031-9155\%2F60\%2F14\%2F5471}

\bibitem{Vallieres2017}
Valli{\`e}res, M., Kay-Rivest, E., Perrin, L.J., Liem, X., Furstoss, C., Aerts,
  H.J.W.L., Khaouam, N., Nguyen-Tan, P.F., Wang, C.S., Sultanem, K., Seuntjens,
  J., El~Naqa, I.: Radiomics strategies for risk assessment of tumour failure
  in head-and-neck cancer.
\newblock Scientific Reports \textbf{7}(1), 10117 (2017).
\newblock \doi{10.1038/s41598-017-10371-5}

\bibitem{Hassani2019}
Hassani, C., Varghese, B.A., Nieva, J., Duddalwar, V.: Radiomics in pulmonary
  lesion imaging.
\newblock American Journal of Roentgenology \textbf{212}(3), 497--504 (2019).
\newblock \doi{10.2214/AJR.18.20623}.
\newblock \urlprefix\url{https://doi.org/10.2214/AJR.18.20623}

\bibitem{Varghese2019}
Varghese, B., Chen, F., Hwang, D., Palmer, S.L., De~Castro~Abreu, A.L.,
  Ukimura, O., Aron, M., Aron, M., Gill, I., Duddalwar, V., Pandey, G.:
  Objective risk stratification of prostate cancer using machine learning and
  radiomics applied to multiparametric magnetic resonance images.
\newblock Scientific Reports \textbf{9}(1), 1570 (2019).
\newblock \doi{10.1038/s41598-018-38381-x}.
\newblock \urlprefix\url{https://doi.org/10.1038/s41598-018-38381-x}

\bibitem{Varghese20192}
Varghese, B.A., Cen, S.Y., Hwang, D.H., Duddalwar, V.A.: Texture analysis of
  imaging: What radiologists need to know.
\newblock American Journal of Roentgenology \textbf{212}(3), 520--528 (2019).
\newblock \doi{10.2214/AJR.18.20624}.
\newblock \urlprefix\url{https://doi.org/10.2214/AJR.18.20624}

\bibitem{Lorensen1987}
Lorensen, W.E., Cline, H.E.: Marching cubes: A high resolution 3d surface
  construction algorithm.
\newblock In: Proceedings of the 14th Annual Conference on Computer Graphics
  and Interactive Techniques, SIGGRAPH ’87, p. 163–169. Association for
  Computing Machinery, New York, NY, USA (1987).
\newblock \doi{10.1145/37401.37422}.
\newblock \urlprefix\url{https://doi.org/10.1145/37401.37422}

\bibitem{NEWMAN2006854}
Newman, T.S., Yi, H.: A survey of the marching cubes algorithm.
\newblock Computers \& Graphics \textbf{30}(5), 854 -- 879 (2006).
\newblock \doi{https://doi.org/10.1016/j.cag.2006.07.021}.
\newblock
  \urlprefix\url{http://www.sciencedirect.com/science/article/pii/S0097849306001336}

\bibitem{Lewiner2003}
Lewiner, T., Lopes, H., Vieira, A.W., Tavares, G.: Efficient implementation of
  marching cubes' cases with topological guarantees.
\newblock Journal of Graphics Tools \textbf{8}(2), 1--15 (2003).
\newblock \doi{10.1080/10867651.2003.10487582}.
\newblock \urlprefix\url{https://doi.org/10.1080/10867651.2003.10487582}

\bibitem{Haralick1973}
{Haralick}, R.M., {Shanmugam}, K., {Dinstein}, I.: Textural features for image
  classification.
\newblock IEEE Transactions on Systems, Man, and Cybernetics \textbf{SMC-3}(6),
  610--621 (1973)

\bibitem{GALLOWAY1975172}
Galloway, M.M.: Texture analysis using gray level run lengths.
\newblock Computer Graphics and Image Processing \textbf{4}(2), 172 -- 179
  (1975).
\newblock \doi{https://doi.org/10.1016/S0146-664X(75)80008-6}.
\newblock
  \urlprefix\url{http://www.sciencedirect.com/science/article/pii/S0146664X75800086}

\bibitem{Thibault2009}
Thibault, G., FERTIL, B., Navarro, C., Pereira, S., Lévy, N., Sequeira, J.,
  MARI, J.L.: Texture indexes and gray level size zone matrix application to
  cell nuclei classification (2009)

\bibitem{Thibault2014}
{Thibault}, G., {Angulo}, J., {Meyer}, F.: Advanced statistical matrices for
  texture characterization: Application to cell classification.
\newblock IEEE Transactions on Biomedical Engineering \textbf{61}(3), 630--637
  (2014)

\bibitem{Amadasun1989}
{Amadasun}, M., {King}, R.: Textural features corresponding to textural
  properties.
\newblock IEEE Transactions on Systems, Man, and Cybernetics \textbf{19}(5),
  1264--1274 (1989)

\bibitem{SUN1983341}
Sun, C., Wee, W.G.: Neighboring gray level dependence matrix for texture
  classification.
\newblock Computer Vision, Graphics, and Image Processing \textbf{23}(3), 341
  -- 352 (1983).
\newblock \doi{https://doi.org/10.1016/0734-189X(83)90032-4}.
\newblock
  \urlprefix\url{http://www.sciencedirect.com/science/article/pii/0734189X83900324}

\bibitem{Mohanaiah2013ImageTF}
Mohanaiah, P., Sathyanarayana, P., GuruKumar, L.: Image texture feature
  extraction using glcm approach.
\newblock International Journal of Scientific and Research Publications
  \textbf{3}(5) (2013)

\bibitem{Humeau2019}
{Humeau-Heurtier}, A.: Texture feature extraction methods: A survey.
\newblock IEEE Access \textbf{7}, 8975--9000 (2019)

\bibitem{Gade2018}
Gade, A.A., Vyavahare, A.J.: Feature extraction using glcm for dietary
  assessment application.
\newblock International Journal Multimedia and Image Processing (IJMIP)
  \textbf{8}(2), 409--413 (2018)

\bibitem{Chernikov2014}
Chernikov, A.N., Xu, J.: A computer-assisted proof of correctness of a marching
  cubes algorithm.
\newblock In: J.~Sarrate, M.~Staten (eds.) Proceedings of the 22nd
  International Meshing Roundtable, pp. 505--523. Springer International
  Publishing, Cham (2014)

\bibitem{DELIBASIS2001343}
Delibasis, K., Matsopoulos, G., Mouravliansky, N., Nikita, K.: A novel and
  efficient implementation of the marching cubes algorithm.
\newblock Computerized Medical Imaging and Graphics \textbf{25}(4), 343 -- 352
  (2001).
\newblock \doi{https://doi.org/10.1016/S0895-6111(00)00082-3}.
\newblock
  \urlprefix\url{http://www.sciencedirect.com/science/article/pii/S0895611100000823}

\bibitem{RAJON2003411}
Rajon, D., Bolch, W.: Marching cube algorithm: review and trilinear
  interpolation adaptation for image-based dosimetric models.
\newblock Computerized Medical Imaging and Graphics \textbf{27}(5), 411 -- 435
  (2003).
\newblock \doi{https://doi.org/10.1016/S0895-6111(03)00032-6}.
\newblock
  \urlprefix\url{http://www.sciencedirect.com/science/article/pii/S0895611103000326}

\bibitem{Parmar2015}
Parmar, C., Leijenaar, R.T.H., Grossmann, P., Rios~Velazquez, E., Bussink, J.,
  Rietveld, D., Rietbergen, M.M., Haibe-Kains, B., Lambin, P., Aerts, H.J.:
  Radiomic feature clusters and prognostic signatures specific for lung and
  head {\&} neck cancer.
\newblock Scientific Reports \textbf{5}(1), 11044 (2015).
\newblock \doi{10.1038/srep11044}.
\newblock \urlprefix\url{https://doi.org/10.1038/srep11044}

\bibitem{CUOCOLO2019144}
Cuocolo, R., Stanzione, A., Ponsiglione, A., Romeo, V., Verde, F., Creta, M.,
  Rocca], R.L., Longo, N., Pace, L., Imbriaco, M.: Clinically significant
  prostate cancer detection on mri: A radiomic shape features study.
\newblock European Journal of Radiology \textbf{116}, 144 -- 149 (2019).
\newblock \doi{https://doi.org/10.1016/j.ejrad.2019.05.006}.
\newblock
  \urlprefix\url{http://www.sciencedirect.com/science/article/pii/S0720048X19301664}

\bibitem{Zhu2015}
Zhu, Y., Li, H., Guo, W., Drukker, K., Lan, L., Giger, M.L., Ji, Y.:
  Deciphering genomic underpinnings of quantitative mri-based radiomic
  phenotypes of invasive breast carcinoma.
\newblock Scientific Reports \textbf{5}(1), 17787 (2015).
\newblock \doi{10.1038/srep17787}.
\newblock \urlprefix\url{https://doi.org/10.1038/srep17787}

\bibitem{Varnas2004}
Varnäs, K., Halldin, C., Hall, H.: Autoradiographic distribution of serotonin
  transporters and receptor subtypes in human brain.
\newblock Human Brain Mapping \textbf{22}(3), 246--260 (2004).
\newblock \doi{10.1002/hbm.20035}.
\newblock
  \urlprefix\url{https://onlinelibrary.wiley.com/doi/abs/10.1002/hbm.20035}

\bibitem{THIBAULT2013}
THIBAULT, G., FERTIL, B., NAVARRO, C., PEREIRA, S., CAU, P., LEVY, N.,
  SEQUEIRA, J., MARI, J.L.: Shape and texture indexes application to cell
  nuclei classification.
\newblock International Journal of Pattern Recognition and Artificial
  Intelligence \textbf{27}(01), 1357002 (2013).
\newblock \doi{10.1142/S0218001413570024}.
\newblock \urlprefix\url{https://doi.org/10.1142/S0218001413570024}

\bibitem{Wahl2009}
Wahl, R.L., Jacene, H., Kasamon, Y., Lodge, M.A.: From recist to percist:
  Evolving considerations for pet response criteria in solid tumors.
\newblock Journal of nuclear medicine : official publication, Society of
  Nuclear Medicine \textbf{50 Suppl 1}(Suppl 1), 122S--50S (2009).
\newblock \doi{10.2967/jnumed.108.057307}.
\newblock \urlprefix\url{https://doi.org/10.2967/jnumed.108.057307}

\bibitem{Frings2014}
Frings, V., van Velden, F.H.P., Velasquez, L.M., Hayes, W., van~de Ven, P.M.,
  Hoekstra, O.S., Boellaard, R.: Repeatability of metabolically active tumor
  volume measurements with fdg pet/ct in advanced gastrointestinal
  malignancies: A multicenter study.
\newblock Radiology \textbf{273}(2), 539--548 (2014).
\newblock \doi{10.1148/radiol.14132807}.
\newblock \urlprefix\url{https://doi.org/10.1148/radiol.14132807}.
\newblock PMID: 24865311

\bibitem{Macyszyn2015}
Macyszyn, L., Akbari, H., Pisapia, J.M., Da, X., Attiah, M., Pigrish, V., Bi,
  Y., Pal, S., Davuluri, R.V., Roccograndi, L., Dahmane, N., Martinez-Lage, M.,
  Biros, G., Wolf, R.L., Bilello, M., O'Rourke, D.M., Davatzikos, C.: {Imaging
  patterns predict patient survival and molecular subtype in glioblastoma via
  machine learning techniques}.
\newblock Neuro-Oncology \textbf{18}(3), 417--425 (2015).
\newblock \doi{10.1093/neuonc/nov127}.
\newblock \urlprefix\url{https://doi.org/10.1093/neuonc/nov127}

\bibitem{ELNAQA20091162}
Naqa], I.E., Grigsby, P., Apte, A., Kidd, E., Donnelly, E., Khullar, D.,
  Chaudhari, S., Yang, D., Schmitt, M., Laforest, R., Thorstad, W., Deasy, J.:
  Exploring feature-based approaches in pet images for predicting cancer
  treatment outcomes.
\newblock Pattern Recognition \textbf{42}(6), 1162 -- 1171 (2009).
\newblock \doi{https://doi.org/10.1016/j.patcog.2008.08.011}.
\newblock
  \urlprefix\url{http://www.sciencedirect.com/science/article/pii/S0031320308003294}.
\newblock Digital Image Processing and Pattern Recognition Techniques for the
  Detection of Cancer

\bibitem{Unser1986}
{Unser}, M.: Sum and difference histograms for texture classification.
\newblock IEEE Transactions on Pattern Analysis and Machine Intelligence
  \textbf{PAMI-8}(1), 118--125 (1986)

\bibitem{Xu2004}
Xu, D.H., Kurani, A., Furst, J., Raicu, D.: Run-length encoding for volumetric
  texture.
\newblock The 4th IASTED International Conference on Visualization, Imaging,
  and Image Processing  (2004)

\bibitem{CHU1990415}
Chu, A., Sehgal, C., Greenleaf, J.: Use of gray value distribution of run
  lengths for texture analysis.
\newblock Pattern Recognition Letters \textbf{11}(6), 415 -- 419 (1990).
\newblock \doi{https://doi.org/10.1016/0167-8655(90)90112-F}.
\newblock
  \urlprefix\url{http://www.sciencedirect.com/science/article/pii/016786559090112F}

\bibitem{Tang1998}
{Xiaoou Tang}: Texture information in run-length matrices.
\newblock IEEE Transactions on Image Processing \textbf{7}(11), 1602--1609
  (1998)

\bibitem{Tustison2009}
Tustison, N., Gee, J.: Run-length matrices for texture analysis.
\newblock The Insight Journal pp. 1--6 (2008)

\end{thebibliography}
\end{document}